\documentclass{article}

\PassOptionsToPackage{numbers, compress}{natbib}
\usepackage[preprint]{neurips_2025}

\usepackage[utf8]{inputenc} 
\usepackage[T1]{fontenc}    
\usepackage{hyperref}       
\usepackage{url}            
\usepackage{booktabs}       
\usepackage{amsfonts}       
\usepackage{nicefrac}       
\usepackage{microtype}      
\usepackage[dvipsnames]{xcolor}         
\usepackage{amsmath}
\usepackage{amssymb}

\usepackage{tikz}
\usetikzlibrary{tikzmark}

\title{Logit-Based Losses Limit the Effectiveness of \\ Feature Knowledge Distillation}

\author{Nicholas Cooper,  Lijun Chen,  Sailesh Dwivedy,  Danna Gurari\\
	University of Colorado Boulder}

\begin{document}
\maketitle

\begin{abstract}
Knowledge distillation (KD) methods can transfer knowledge of a parameter-heavy teacher model to a light-weight student model. The status quo for feature KD methods is to utilize loss functions based on logits (i.e., pre-softmax class scores) and intermediate layer features (i.e., latent representations).  Unlike previous approaches, we propose a feature KD framework for training the student's backbone using feature-based losses \emph{exclusively} (i.e., without logit-based losses such as cross entropy). Leveraging recent discoveries about the geometry of latent representations, we introduce a \emph{knowledge quality metric} for identifying which teacher layers provide the most effective knowledge for distillation. Experiments on three image classification datasets with four diverse student-teacher pairs, spanning convolutional neural networks and vision transformers, demonstrate our KD method achieves state-of-the-art performance, delivering top-1 accuracy boosts of up to $15\%$ over standard approaches. We publicly share our code to facilitate future work at this \href{https://github.com/Thegolfingocto/KD_wo_CE.git}{GitHub page}.
\end{abstract}

\section{Introduction}
Knowledge distillation (KD) is a popular approach for model compression which infuses ``dark knowledge'' from a parameter-heavy teacher model into a more compact student model. Fundamental to KD is the question: \emph{What knowledge should be transferred}? Prior work proposes two answers. \emph{First}, the pioneering KD paper~\cite{SKD} standardized that a student should be trained to mimic the teacher's softened class probability distribution (i.e., temperature-scaled softmax).  This is accomplished by regularizing cross entropy (CE) loss with the KL divergence between the softened logits of the two models, making this method solely reliant on logit-based supervision.  \emph{Second}, FitNets~\cite{FitNets} pioneered the popular trend of using additional loss terms to guide the student to also mimic the teacher’s intermediate layer features (i.e., latent representations), by minimizing the distance between their corresponding feature maps.  We refer to these two techniques as \emph{vanilla knowledge distillation} (VKD) and \emph{feature knowledge distillation} (FKD), respectively.

Our work is motivated by the hypothesis that \emph{the performance of FKD methods is compromised by training student backbones with logit-based losses.}   This is inspired by the observation that there are limits to the information logit-losses can transfer.  Feature-based losses are computed in very high dimensional spaces, enabling them to capture richer information about a teacher's representations.  Logit-based losses, in contrast, are computed in lower dimensional spaces, limiting the level of detailed knowledge they can transfer.  We suspect that such lower-dimensional losses dilute the potential effectiveness of FKD methods.

\begin{figure}
	\includegraphics[width=0.785\textwidth]{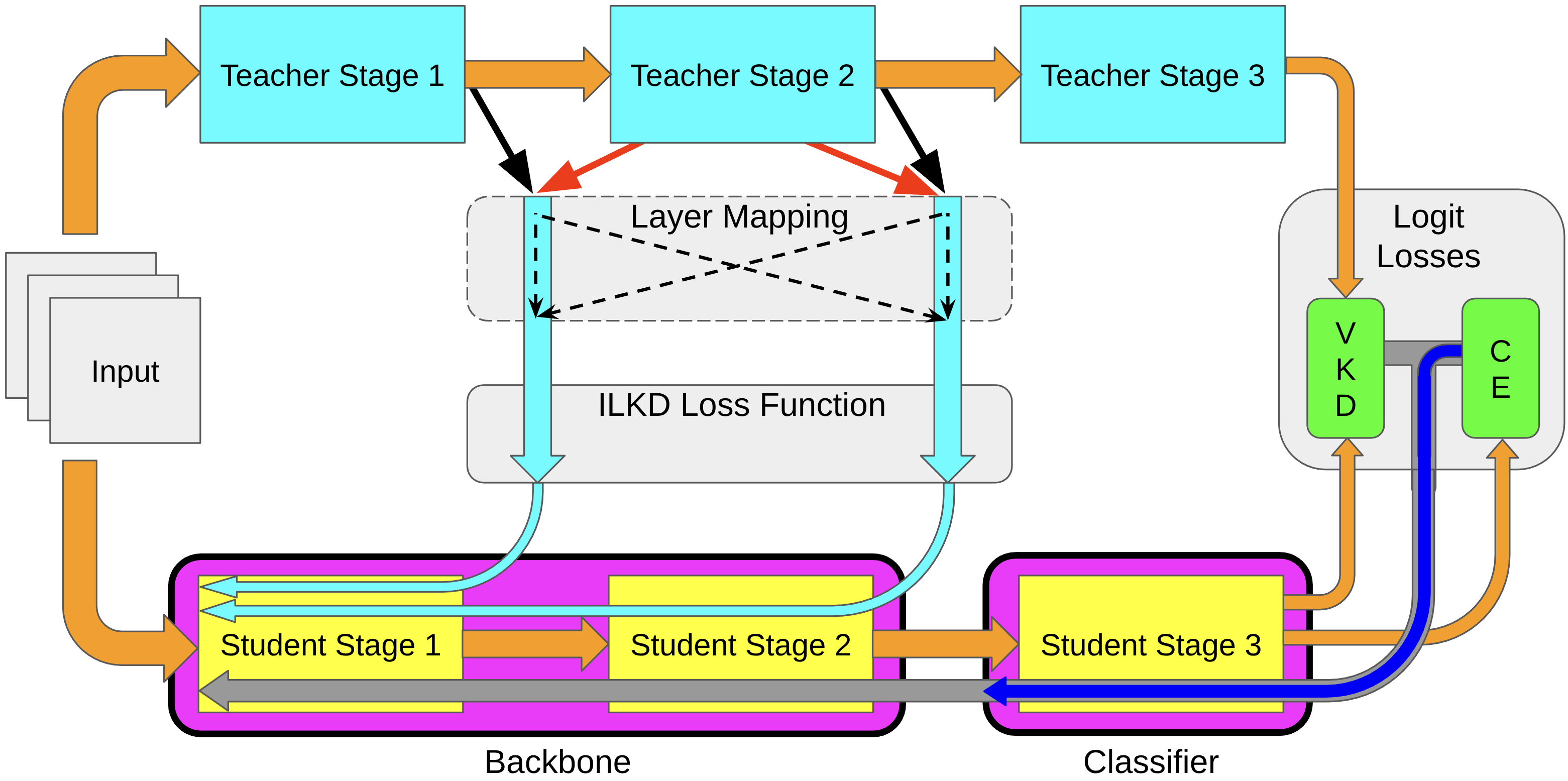}
	\includegraphics[width=0.20\textwidth]{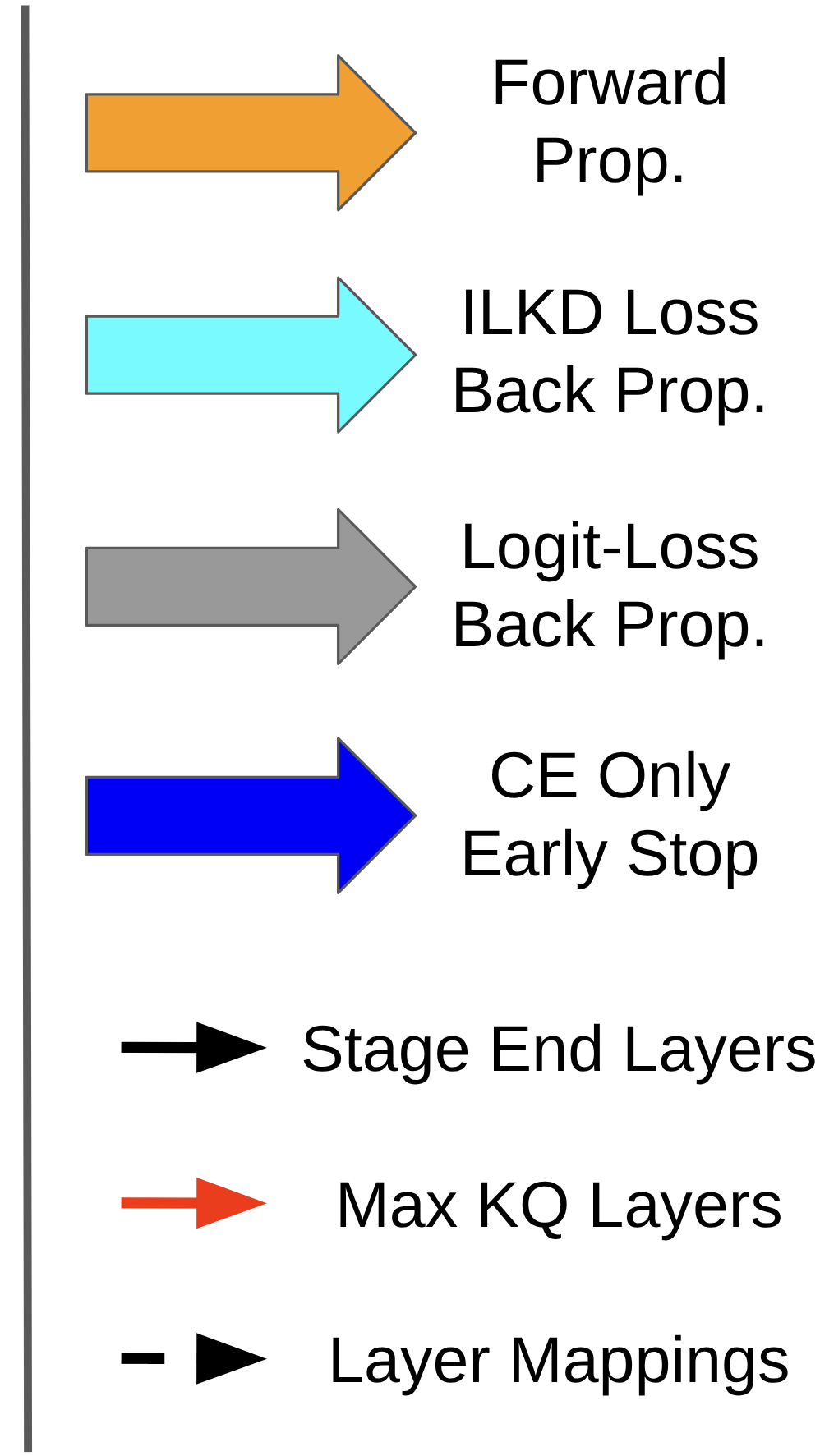}
	\caption{Illustration of our feature knowledge distillation framework and its two key distinctions from prior work.  First, while the status quo is to back-propagate logit-based losses through the student backbone ({\color{gray}$\rightarrow$}), our method back-propagates only CE through the just the classifier ({\color{blue}$\rightarrow$}).  Second, while the default strategy is to select teacher layers to distill from the end of each `stage' ({\color{black}$\rightarrow$}), we introduce a metric for automatically selecting the layers with the highest knowledge quality which often occur within a single stage ({\color{red}$\rightarrow$}).}
    \label{fig:MethodOverview}
\end{figure}

Accordingly, we propose a new FKD framework for training student backbones \emph{only} with feature-based losses (i.e., no logit-based loss terms), as illustrated in \textbf{Figure \ref{fig:MethodOverview}}. Here, the \emph{student backbone} refers to the layers which receive information from the teacher's intermediate layers. Our key insight is that this loss recipe often fails when standard techniques are used to select intermediate teacher layers for distillation (\textbf{Figure \ref{Experiment2}}).  We introduce a novel, geometry-aware layer selection metric to automatically select intermediate teacher layers with the highest \emph{knowledge quality} (KQ) for distillation. Experiments with three image classification datasets and four diverse student-teacher pairs, spanning convolutional neural networks (CNNs) and vision transformers (ViT), demonstrate that our proposed approach outperforms existing KD baselines by up to $15\%$ top-1 accuracy. Our fine-grained analyses reveal that (1) logit-based losses prevent the student from fully benefiting from distillation and (2) our teacher layer selection method is necessary to guarantee a performance boost when excluding logit-based losses.

\section{Related Work}

\paragraph{Feature Knowledge Distillation (FKD).}
FKD involves three intertwined steps: (1) teacher \textit{layer selection}, (2) teacher-to-student \textit{layer mapping}, and (3) teacher-to-student \textit{dimensional translation} to match the teacher's higher dimensional latent space to that of the student.  While much research focuses on the second step~\cite{TeacherAttnKD,InfMatchKD, StudentAlignedKD} and third step~\cite{FitNets, TeacherAttnKD, InfMatchKD, DimReductionKDGood, ProjGood, AT, TargetAwareTransformerKD, XArchKD, ReusedClsKD, OFAKD, RKD, IRG, SPKD, TLRGKD, GNNKD}, the initial step of teacher layer selection has received little attention.  Most works default to selecting the final layer from each teacher model's \emph{stage}, defined as layers preceding a pooling operation.  We instead introduce the first automated metric for teacher layer selection, which identifies layers with the highest knowledge quality.

\vspace{-0.5em}\paragraph{Distillation Without Logit Losses.}
Several works have hinted at potential benefits of training the student backbone without logit losses. For example, the classic FitNets \cite{FitNets} excluded logit losses during the first part of a two-stage process, but then introduced them in the second phase. More recently, \cite{TargetAwareTransformerKD} showed that removing the VKD loss term does not significantly affect performance. Most similar to our work is \cite{StudentAlignedKD}, which demonstrated that FKD methods suffer from mis-alignment between gradients of the logit and feature losses.  However, they addressed this issue by dynamically disabling the feature loss rather than removing logit losses. Complementing these works, we instead propose distilling student backbones \emph{without any} logit losses. Fine-grained analysis validates it is most beneficial to remove \emph{both} logit loss terms, rather than only the VKD loss term removed in \cite{TargetAwareTransformerKD}.

\vspace{-0.5em}\paragraph{Geometry of Latent Representations.}
Recent observations about the latent geometry of image classification models inspired the design of our metric for measuring the knowledge quality of teacher layers. In particular, growing evidence shows that models process data in two distinct ways: \emph{extraction} and \emph{compression} \cite{TunnelEffect}. The notion of \emph{compression} first emerged with Deep Neural Collapse (DNC) \cite{SemNC}, which demonstrated that models learn to \emph{compress} same-class points towards the class mean in the final layer representations, and was later extended to latent representations~\cite{IntermediateNeuralCollapse}. The notion of \emph{extraction} emerged from observations about the \textit{intrinsic dimension} of latent representations~\cite{RandomIDPerLayer, IDStagesNNs, IDPerLayer, RegularizationAndGenVsID}, showing that image classification models process data by \textit{expanding representation dimensionality} in their early layers (e.g., the first $~70\%$ of layers) before \textit{compressing representation dimensionality} in later layers (e.g., the final $~30\%$ of layers). Our proposed metric is based on three geometric properties of a teacher's feature representations,\footnote{While prior work~\cite{ManifoldKD} explored latent feature geometry to improve FKD, they focused on the \emph{dimensional translation} step rather than teacher \emph{layer selection}.}.  Experiments reveal that the \emph{transition layers} between extraction and compression exhibit the highest `knowledge quality' and yield the best KD performance.
\section{Methods}

\subsection{Basic Notation and Background} 
We consider teacher and student models as sequences of parameterized functions $f(x) = f_l \circ ... f_2 \circ f_1(x): \mathbb{R}^d \rightarrow \mathbb{R}^C$, where $d$ is the dimensionality of the input, $C$ the number of class labels, and $l$ the number of layers. The training dataset $(X, Y)$ consists of a set $X \in \mathbb{R}^{N\times d}$ of $N$ $d$-dimensional inputs and a set $Y \in \{1, 2, ..., C\}^N$ of ground truth labels. 

We use the term \textit{representation} to refer to the output from a sub-sequence of layers. For example, the representations from the $i^{th}$ layer, denoted by $R_i$, is the set: 
\begin{equation*}
    R_i = \{f_i \circ f_{i-1} \circ ...f_2(x) \circ f_1(x) \textbf{ } | \textbf{ } x \in X\} \subset \mathbb{R}^{d_i}
\end{equation*}

\noindent
where $d_i$ denotes the \textit{ambient dimension} of the $i^{th}$ layer's representations, e.g., a linear layer with $100$ neurons has an ambient dimension of $100$. Putting this together, models can be written in the following form:
\begin{equation*}
    f(x): \mathbb{R}^d \xrightarrow{f_1} \mathbb{R}^{d_1}  \xrightarrow{f_2} \mathbb{R}^{d_2}...\mathbb{R}^{d_{l-1}} \xrightarrow{f_{l}} \mathbb{R}^C
\end{equation*} 
\noindent
Note that we do not consider architecture-specific tensor structure (e.g., CNN feature maps with $c$ channels worth of $h\times w$ sized features are interpreted as ``flattened'' vectors in $\mathbb{R}^{chw}$).  The final layer's representations, $R_l \subset \mathbb{R}^C$, are called \textit{logits}.

For layer selection, we denote the index sets of the selected teacher and student layers by $L^T \subseteq \{ 1, 2, ..., l^T\}$ and $L^S \subseteq \{ 1, 2, ..., l^S\}$, where $l^T$ and $l^S$ denote the total number of teacher and student layers respectively. The mapping between the teacher and student layers is then defined by a {\em mapping matrix} $A\in \mathbb{R}^{|L^T|\times|L^S|}$ that assigns weights to the losses computed between each of the teacher-student layer pairs. These weights can be dynamically learned \cite{TeacherAttnKD} or adjusted via handcrafted rules \cite{InfMatchKD, StudentAlignedKD}.

FKD learning involves two types of the losses.  First is the classic VKD loss \cite{SKD}, computed as follows:
\vspace{-0.5em}
{\small \begin{equation}
    \mathcal{L}_{KL}(R^S_{n}, R^T_{n}, t) = t^2KL \left[ softmax \left( \frac{R^S_{n}}{t} \right), softmax \left( \frac{R^T_{n}}{t} \right) \right]
    \label{VKDLoss}
\end{equation}}
\noindent
where $KL[\cdot, \cdot]$ denotes the KL divergence, $softmax(\cdot)$ the softmax function, $t$ the temperature, and $t^2$ a balancing constant that counteracts the decay of $KL$ when logits are softened.  Second is a loss computed between pairs of teacher-student intermediate layers, resulting in the following total loss:
{\small \begin{equation}
	\mathcal{L}_{FKD} = \mathcal{L}_{CE} + \mathcal{L}_{KL} + \sum_{i \in L^T}\sum_{j \in L^S}A_{ij}\mathcal{L}_{F}(R^T_i, R^S_j)
    \label{FKDLoss}
\end{equation}}
\noindent
where $\mathcal{L}_{CE}$ denotes cross-entropy loss and $\mathcal{L}_{F}$ denotes the feature loss computed after aligning the teacher and student ambient dimensions.

\subsection{Proposed Method}
In what follows, we formalize the two complementary innovations of our proposed FKD framework, illustrated in \textbf{Figure~\ref{fig:MethodOverview}}, and describe our implementation.

First, we modify the standard loss recipe in \textbf{Equation \ref{FKDLoss}} by removing \textit{all} logit losses from the student backbone during training. Here ``backbone'' is defined as the set of layers $\{ f^S_i \text{ } | \text{ } i \leq l^S_{final} \}$, where $l^S_{final} := max(L^S)$ denotes the last (deepest) student layer selected for distillation. We drop the $\mathcal{L}_{KL}$ term completely, and stop back-propagation of $\mathcal{L}_{CE}$ at $l^S_{final}$. This means the student backbone is trained to minimize only the feature-based loss, $\mathcal{L}_{F}$, while the student classifier is trained to minimize only the cross-entropy loss $\mathcal{L}_{CE}$.

Second, for teacher layer selection (i.e, $L^T$), we choose the top-$k$ layers with the highest knowledge quality from our metric, which we denote by $\mathcal{Q}$.  This metric combines three geometric properties of layer representations as follows:
\vspace{-0.5em}
{\small
\begin{equation}
	\mathcal{Q}(R) := \mathcal{S}(R) + \sqrt{\mathcal{I}(R)\mathcal{E}(R)}
    \label{FQ}
\end{equation}
}
\hspace{-1mm}
the components being measures of \textit{separation} ($\mathcal{S}$), \textit{information} ($\mathcal{I}$), and \textit{efficiency} ($\mathcal{E}$).  These are computed from the average within-class dot product (avgDPW), between-class dot product (avgDPB), minimum within-class dot product (minDPW), between-class distance (minDistB), and average norm (avgNorm).  For a set of representations $R=\{r_1, r_2, \cdots, r_N\}$ of the training dataset, these are defined as follows:
{\small
\begin{equation*}
	\begin{aligned}
		avgDPW (R)&= \frac{1}{C} \sum_{c=1}^C\sum_{(i, j)\in N_c^2}\frac{\langle r_i \cdot r_j\rangle}{|N_c|^2-|N_c|}
        \\
		avgDPB (R)&= \frac{2}{C^2-C} \sum_{c=1}^C \sum_{c'=c+1}^C\sum_{i\in N_c}\sum_{j\in N_{c'}}\frac{\langle r_i \cdot r_j\rangle}{|N_c||N_{c'}|}
        \\
		minDPW (R)&= \frac{1}{C} \sum_{c=1}^C \min_{(i, j)\in N_c^2}|\langle r_i \cdot r_j\rangle|
        \\
            minDistB (R) &= \frac{2}{C^2-C} \sum_{c=1}^C \sum_{c'=c+1}^C\min_{i\in N_c, j\in N_{c'}}\|r_i-r_j\|_2\\
            avgNorm (R)& = \frac{1}{N} \sum_{i=1}^N \|r_i\|_2
	\end{aligned}
\end{equation*}
}
\hspace{-1mm}where $\langle r_i \cdot r_i\rangle= \frac{r_i\cdot r_j}{\|r_i\|_2\|r_j\|_2}$ is the normalized dot product (i.e., cosine similarity), $N_c$ is the subset of indices corresponding to class $c$,  and $(i, j)\in N_c^2$ denotes a distinct pair of such indices.

\emph{Separation} measures the extent to which a representation distinguishes classes by using the average within-class cosine similarity and between-class cosine similarity:
{\small
\begin{equation}
	\mathcal{S}(R) = avgDPW(R) - avgDPB(R)
        \label{SDef}
\end{equation}
}
\noindent
\hspace{-1mm}Prioritizing layers with higher scores ensures the selected teacher layers can convey rich information about class labels to the student.  As we will show in the experiments, separation tends to increase during the compression layers of the model, which is consistent with prior findings \cite{IntermediateNeuralCollapse}.

\emph{Information} measures the richness of a layer's ``dark knowledge" by considering its embedding dimension and within-class variation.  Our use of the \textit{embedding dimension} aligns with prior work~\cite{RegularizationAndGenVsID}, which found that peak intrinsic dimension (ID)---i.e., the maximum ID achieved across all layers---is correlated with better model performance.  While in practice it can be difficult to measure the intrinsic dimension---specifically, the \emph{actual} dimension of the latent feature manifold---from samples (see \cite{IDSurvey, M2ID, TwoNNID}), the embedding dimension serves as a reasonable proxy that is relatively easy to compute (e.g., with principal component analysis).  We also incorporate within-class variance since the ground truth labels lack such information. Formally, we combine the minimum within-class similarity (a proxy for within-class variance), and the average class-wise normalized \textit{SVD Entropy} (a measure of embedding dimension) as follows:
\begin{equation}
        \mathcal{I}(R) = \left[1 - minDPW(R)\right]  avgSVDE(R)
    \label{IDef}
\end{equation}
where $avgSVDE(\cdot)$ is defined as:
\vspace{-1em}
\begin{equation*}
    avgSVDE(R) = \frac{1}{C}\sum_{c=1}^C\frac{H(\bar{\mathbf{\sigma_c}})}{ln(N_c)}
\end{equation*}
\noindent
where $H(\bar{\mathbf{\sigma}}_c) = -\sum_{i=1}^{d_c}\bar{\mathbf{\sigma}}_{ci} ln(\bar{\mathbf{\sigma}}_{ci})$ is the Shannon Entropy of the normalized singular values ($\bar{\mathbf{\sigma}}_{c}$) of the class covariance matrix and $d_c$ is the embedding dimension of class $c$ estimated by PCA \footnote{In our experiments, we use the number of principal components required to account for $95\%$ of the variance as an estimate of embedding dimension.}. $H(\bar{\mathbf{\sigma}}_c)$ attains a maximum value of $ln(N_c)$ when all $N_c$ data points are equidistant from the mean and mutually pair-wise orthogonal, so we normalize it to get values between 0 and 1. Intuitively, $avgSVDE$ measures the dimensionality of the representations weighted by how evenly their variance is distributed across dimensions.  We will show experimentally that models increase $\mathcal{I}$ during the extraction layers and then decrease $\mathcal{I}$ in order to increase $\mathcal{S}$ during the compression layers. 

\emph{Efficiency} measures how large the representations are relative to how large they must be to allow their information ($\mathcal{I}$). This is motivated by our empirical observation that representations of large norm often lead to student training instability, which we suspect results from the corresponding increase in the magnitude of $\mathcal{L}_{F}$. To formalize this, we consider a hyper-spherical packing problem. Suppose there are $N$ data points and a minimum distance of $\varepsilon$ is required between any two of them, i.e., $||r_i - r_j||_2 \geq \varepsilon,  \forall i\neq j$. Then, for ReLU-family (e.g., ReLU, Leaky ReLU, GELU) activated $D$-dimensional Euclidean space, the radius of the smallest hypersphere that can accommodate such points can be approximated by:
\begin{equation}\label{DimDistTradeOff}
        r_{min}(\varepsilon, d) \approx  \text{ } 2K\varepsilon
\end{equation}
with $K =\left( {\frac{N}{\pi}} \right) ^ {\frac{1}{D-1}}$. \textbf{Equation \ref{DimDistTradeOff}} describes how dimension and radius affect the number of available ``$\varepsilon$-rooms'' on a hypersphere's surface. We set $\varepsilon = minDistB(R)$, then define the packing efficiency as the ratio between the empirical norm and the estimate of the smallest required norm:
{\small
\begin{equation}
	\mathcal{E}(R) = \frac{2KminDistB(R)}{avgNorm(R)}.
\end{equation}
}
\noindent
\hspace{-1mm}Consistent with $\mathcal{I}$, we use a PCA estimate of the global embedding dimension for $D$. $\mathcal{E}$ signifies when representations are size efficient, with lower values indicating the norm is unnecessarily inflated. This value facilitates selecting teacher layers which will improve the student's ability to converge during training. As we will show in the experiments, $\mathcal{E}$ resembles $\mathcal{I}$ in that it increases during the extraction layers, then decreases during compression. 

We make several \emph{implementation} choices for our proposed KD method. We follow existing work \cite{TeacherAttnKD, ReusedClsKD} and set $|L^T| = |L^S| = 4$ with a simple one-to-one layer mapping. We only consider representations of the training dataset $X$ when computing $\mathcal{Q}$ and selecting $L^T$.  We restrict the definition of ``layer'' to minimal sets of functions ending with non-linear activation, since nonlinearities influence the geometric properties of $R_l$,\footnote{For example, ReLU restricts all points to live in the non-negative orthant, the relative volume of which vanishes in high dimension as $\frac{1}{2^d}$.}.  For dimensional translation, we construct a single layer projector. It consists of a pooling layer for spatial alignment, followed by a convolution layer with minimal filters and no bias. We form it directly from the projectors used in \cite{TeacherAttnKD} but reduce the number of layers to 1. In principle, this formulation of $\mathcal{L}_{IL}$ is very similar to \cite{FitNets}. We use a 2D version for CNNs, and a 1D version for transformers. They are identical up to the order of the parameter tensors. We constructed our KQ metric from $\mathcal{S}, \mathcal{I}$, and $\mathcal{E}$ as shown in \textbf{Equation \ref{FQ}} following preliminary empirical analysis that showed this combination of the three components led to the best results.

\vspace{-0.5em}
\section{Experiments}
We now describe our experiments validating improvements from our proposed approach and the importance of its different design choices.

\vspace{-0.5em}\paragraph{Model Architectures.}
To demonstrate that our method is applicable to different architectures, we experiment with convolutional neural networks (CNNs) and vision transformers (ViTs). For \emph{teachers}, we chose VGG19 \cite{VGG}, ResNet34 \cite{ResNet}, and ViT\_B \cite{ViT} to examine the efficacy of our $\mathcal{Q}$ metric on a ``vanilla'' CNN, a residual CNN, and an attention-based architecture.  For \emph{students}, we chose VGG11, MobileNetV2 \cite{MobileNet}, ResNet9, and ViT\_ET (extra tiny) to cover the same three classes of architecture. We include MobileNet because it is a common student choice for KD experiments~\cite{SPKD, TeacherAttnKD, InfMatchKD}. ResNet9 is formed by taking half the layers of ResNet18, another popular student architecture, to create a larger performance gap between teacher and student. Similarly, we use a cut-down vision transformer (ViT\_ET) which has a comparable number of parameters to the other three students.

\vspace{-0.5em}\paragraph{Training Protocol.}
Models are trained for $50$ epochs with the Adam optimizer \cite{Adam} and a single cycle learning rate schedule \cite{OneCycle}. We follow the suggested procedure of \cite{OneCycle} and select a maximum learning rate for each model based on shorter training runs. We train without data augmentation so knowledge quality is measured in a reproducible manner.  Results with data augmentation are provided in the appendices, and reinforce our findings.

\vspace{-0.5em}\paragraph{Datasets.}
We use three datasets commonly employed for knowledge distillation experiments.  Two are CIFAR10 and 100 datasets~\cite{CIFAR}, which are composed of 60,000 32x32 color images from 10 and 100 classes respectively. The third is Tiny ImageNet~\cite{TIM} which consists of 110,000 64x64 color images from 200 classes. Altogether, they reflect a diversity of task difficulties with CIFAR10 a trivial baseline, CIFAR100 more challenging, and Tiny ImageNet that most challenging.

\vspace{-0.5em}\paragraph{Evaluation Metrics.}
We evaluate with top-1 accuracy and average\footnote{Averaged over all relevant baseline distillation methods.} absolute relative improvement (ARI) ~\cite{CRD}.  ARI indicates the benefit of $KD_1$ \textit{relative to} $KD_2$:
\vspace{-0.5em}\begin{equation*}
	ARI(KD_1, KD_2) =  \frac{Acc_{KD_1} - Acc_{KD_2}}{Acc_{KD_2} - Acc_{Baseline}}
\end{equation*}
where $Acc_{Baseline}$ denotes the accuracy of the student model without any distillation. For all experiments, we report mean and standard deviations from three, randomly initialized runs.

\subsection{Analysis: Where Does Peak KQ Occur?}
\begin{figure}[h!]
	\centering
	\includegraphics[width = 0.49\textwidth, height = 3.2cm]{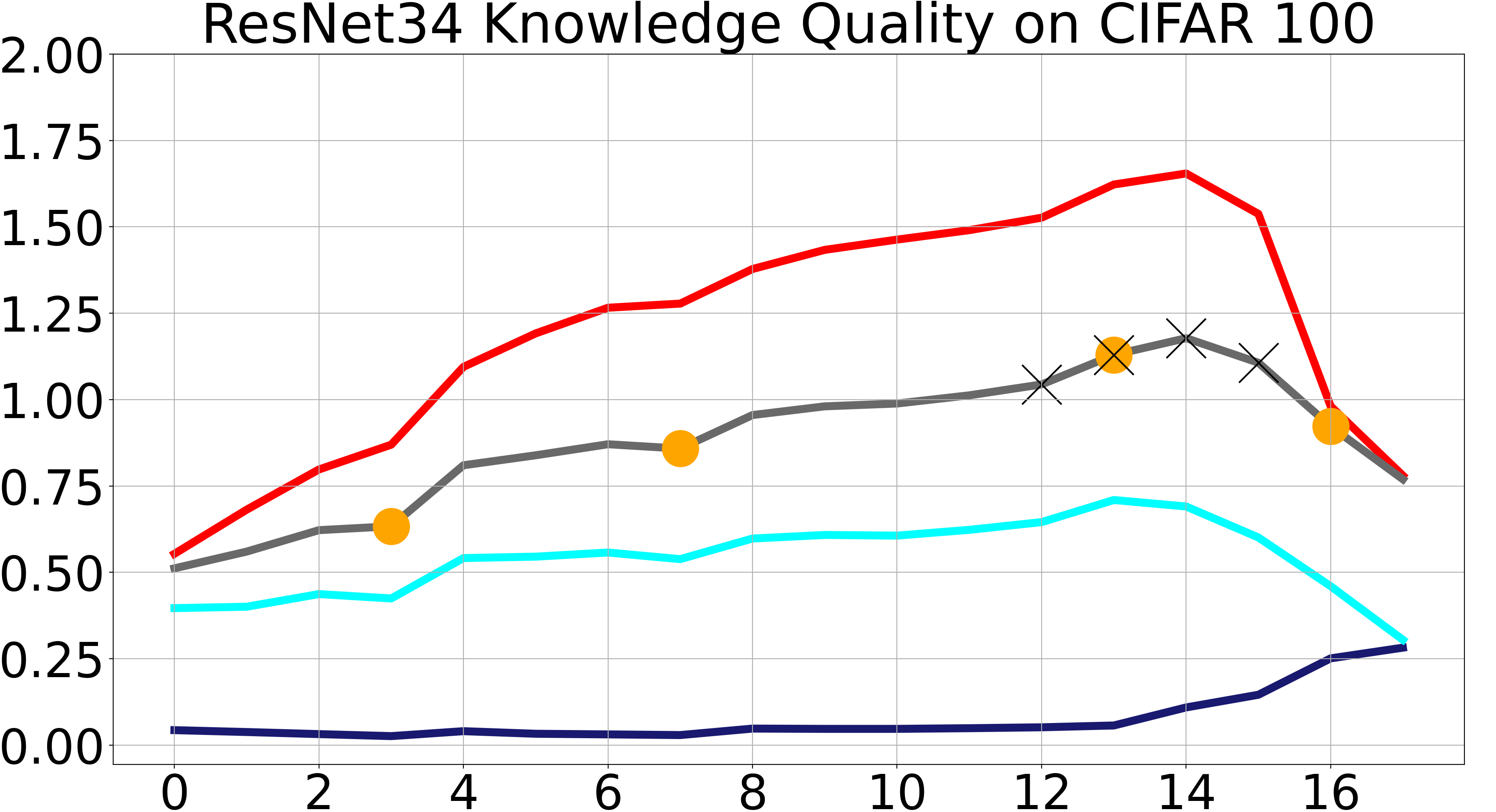}
        \includegraphics[width = 0.49\textwidth, height = 3.2cm]{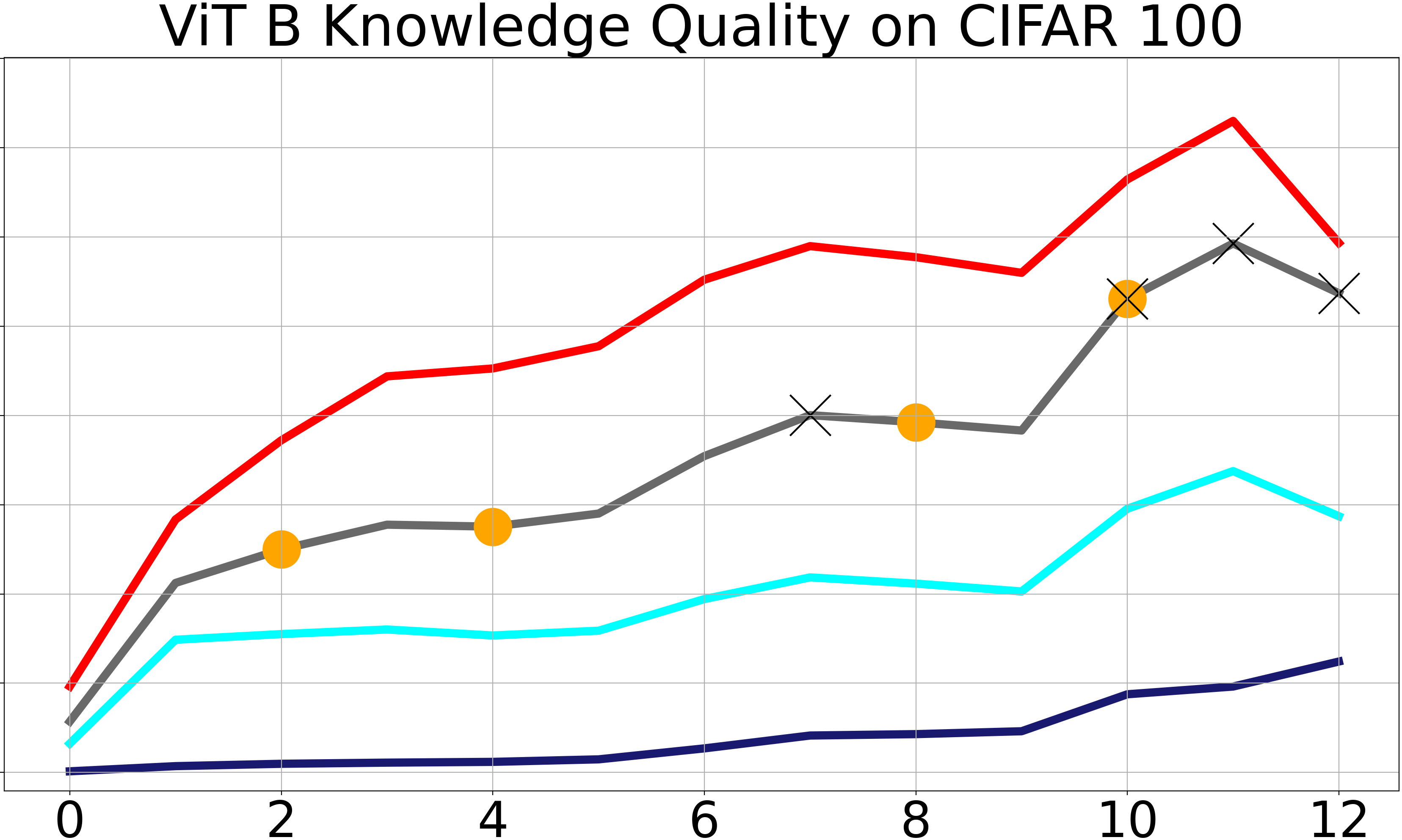}
	\caption{Per-layer knowledge quality analysis of ResNet34 (left) and ViT\_B (right) on CIFAR100. X-axes: layer indices. Y-axis: $\mathcal{S}$ ({\color{NavyBlue}dark blue}), $\mathcal{I}$ ({\color{CornflowerBlue}light blue}), $\mathcal{E}$ ({\color{Red}red}), $\mathcal{Q}$ ({\color{Gray}gray}). {\color{Orange}Orange} circles indicate standard layer selections and {\color{Gray}gray Xs} indicate maximal knowledge quality layers.}
    \label{fig:FQCurves}
\end{figure}

Common trends for the knowledge quality curves are exemplified in \textbf{Figure \ref{fig:FQCurves}} for ResNet34 and ViT\_B on CIFAR100. The components of $\mathcal{Q}$---$\mathcal{S}, \mathcal{I}$, and $\mathcal{E}$---exhibit clear extraction and compression phases in both models, reinforcing prior work's findings \cite{TunnelEffect}.  In the final layers, separation increases at the cost of information and efficiency, a trend that generalizes across datasets (see appendicies).  These findings also show $\mathcal{Q}$ peaks at different relative depths for the models with only ResNet34 exhibiting knowledge quality decay in the final layers. This suggests that ViT\_B contains better knowledge at the logits layer, which we validate experimentally.

\subsection{Distillation Results}

\begin{figure*}[t!]
	\includegraphics[width=\textwidth, height = 8cm]{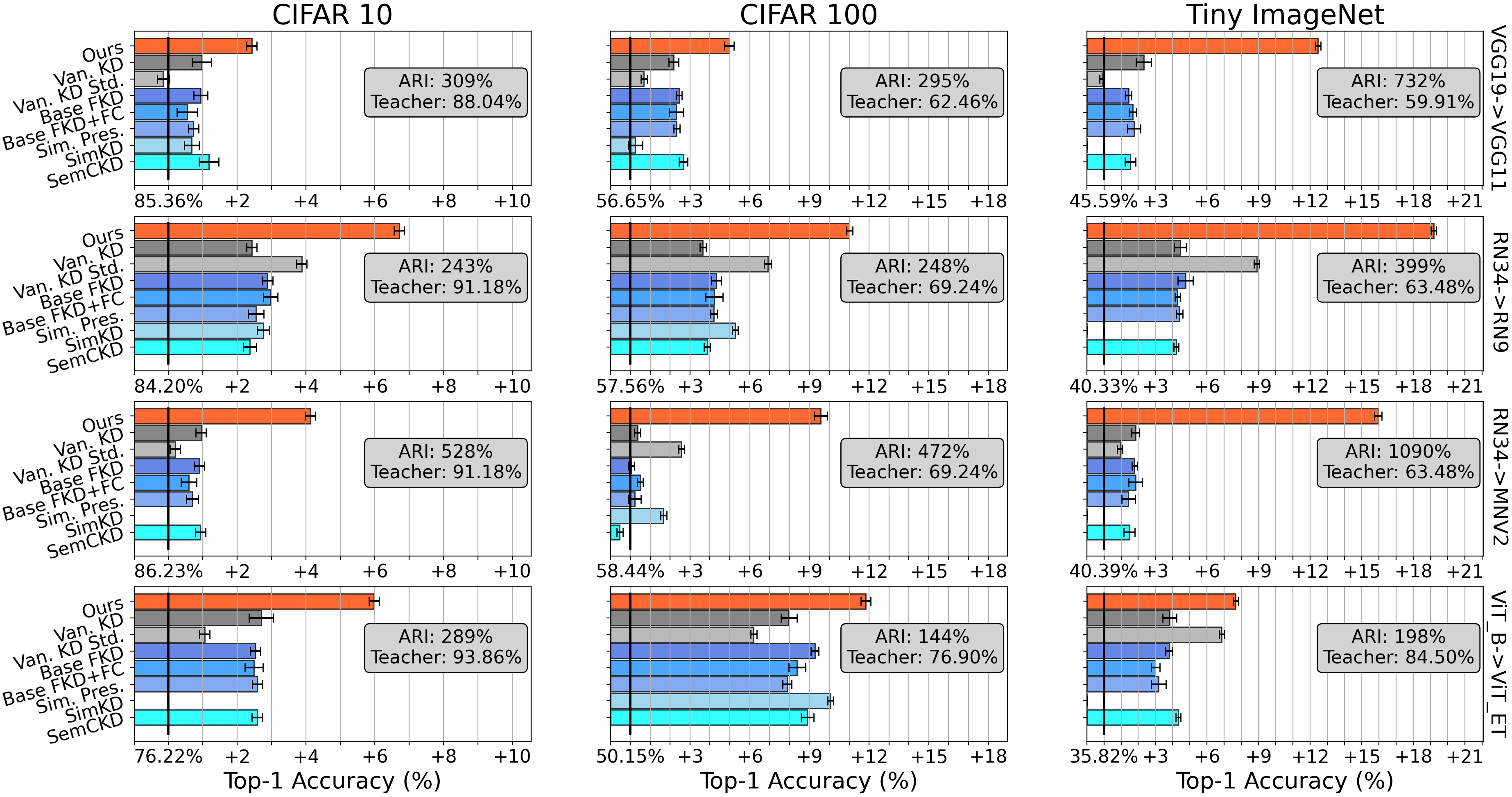}
	\caption{Performance of proposed method and baselines. Vertical black lines denote baseline student performance and the end of each bar shows standard deviation values from three runs.  Configurations which failed to converge are not plotted. ARI denotes the mean ARI from our method to all baselines. }
	\label{Experiment1}
\end{figure*}

\paragraph{Experiment 1: Ours vs. Baselines.}
We compare our FKD method to seven baselines: Vanilla KD (Van. KD \cite{SKD}), Logit Standardization (Van. KD Std. \cite{LogitStandardizationKD}), Base FKD (a modernization of FitNets \cite{FitNets}), Base FKD + FC (base FKD with fully connected layer mapping), Similarity Preserving (Sim. Pres. \cite{SPKD}), SemCKD \cite{TeacherAttnKD}, and Reused Teacher Classifier (SimKD \cite{ReusedClsKD}). These represent top-performing logit and feature KD strategies. For the baselines, we use the widely accepted standard of selecting teacher layers at the end of each model stage which occur next to pooling operations. For ViT, which lacks pooling operations, we select layers mimicking the average relative depths chosen for the CNNs. For baselines which use $\mathcal{L}_{KL}$, we adopt common practice~\cite{TeacherAttnKD, SPKD} and set $t = 4$. Results are shown in \textbf{Figure \ref{Experiment1}}.

Overall, our method considerably outperforms all baselines. Moreover, we observe a positive correlation between dataset difficulty and the performance gains achieved by our method, with the highest ARI scores obtained on Tiny ImageNet. This suggests that the benefit of training the student's backbone without logit losses increases as the classification task becomes less easily separable. We suspect this is because harder datasets result in teacher representations with higher intrinsic dimensionality, thereby offering richer knowledge for the student to learn from.

Examining the influence of model architecture, our method achieves similar absolute benefits but smaller ARIs on transformers than CNNs. We attribute this difference to the way CNNs and ViTs modify knowledge quality across their layers, as shown in \textbf{Figure \ref{fig:FQCurves}}. ViT\_B increases $\mathcal{Q}$ in an almost monotone fashion across its layers, whereas CNNs exhibit degrading knowledge quality in their final layers. As predicted by this analysis, logit loss based methods perform better (by $\sim2x$) on this model pair because ViT contains higher quality knowledge at its final layers than the CNNs.

\begin{figure*}[t!]
	\includegraphics[width=\textwidth, height = 8cm]{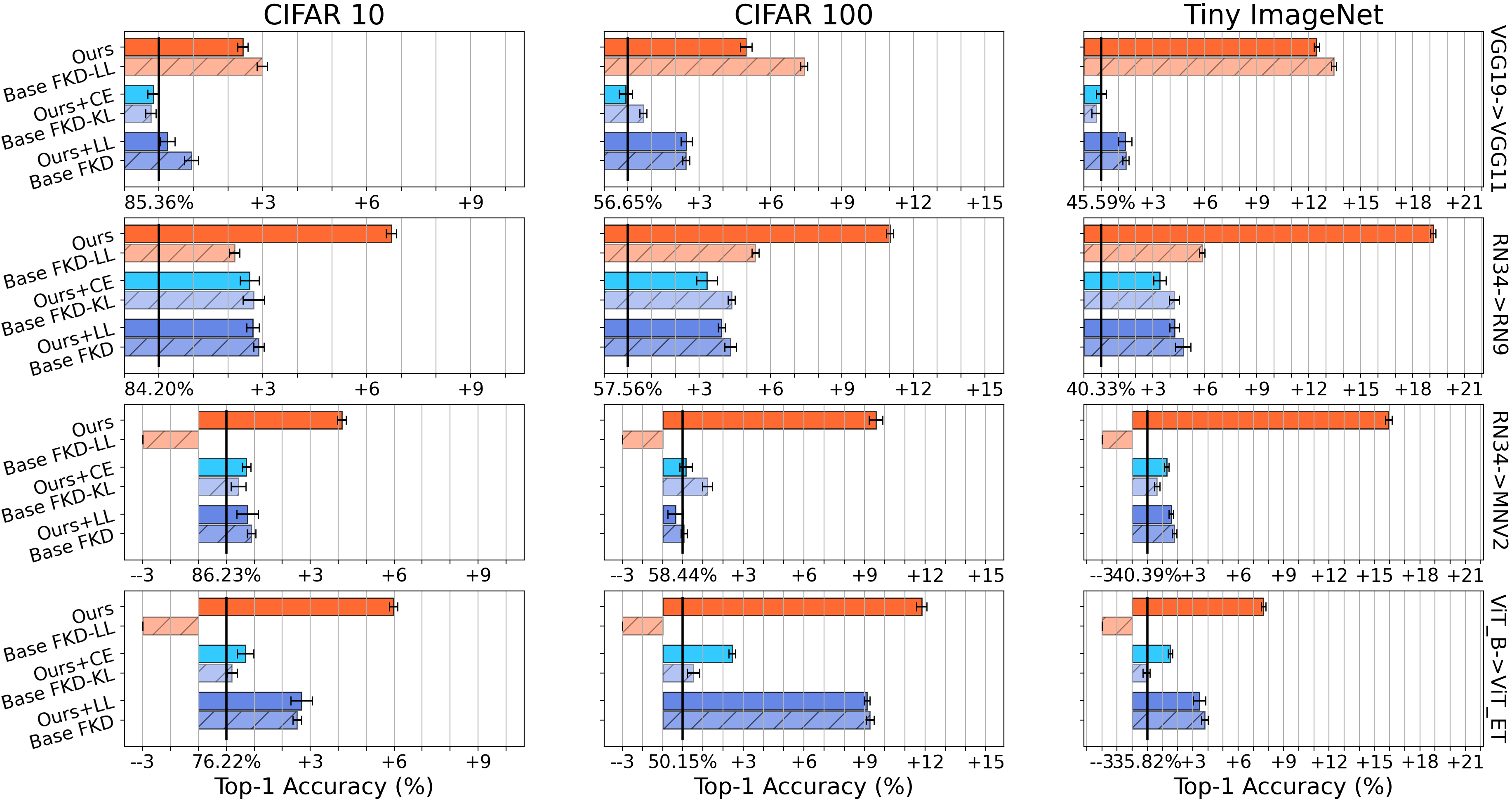}
	\caption{Performance of different teacher layer selection methods when paired with three loss recipes: our loss recipe ({\color{RedOrange} Orange}), CE loss used in backbone ({\color{CornflowerBlue} Light blue}), and both CE and KL loss used in backbone ({\color{NavyBlue} Dark blue}).  Configurations which failed to converge are clipped to $-3$ for improved legibility. Ours and standard layer selection are indicated by solid and striped bars, respectively.}
    \label{Experiment2}
\end{figure*}

\vspace{-0.5em}\paragraph{Experiment 2: When Does $\mathcal{Q}$-Based Layer Selection Matter?}
We next test both our method and the Base FKD baseline (which are identical up to layer selection and loss recipe) with all combinations of layer selection and loss recipe (holding dimensional translation constant). Specifically, for our method, we added both the CE and KL logit losses (``Ours+LL'') and just the CE loss (``Ours+CE''). For Base FKD, we removed the KL loss (``Base FKD-KL'') and all logit losses (``Base FKD-LL'').   Results are in \textbf{Figure \ref{Experiment2}}. We found our layer selection method is critical, as the student failed to converge during training for $6$ of $12$ cases using standard layer selection. When logit losses \textit{are} used, our layer selection method has a smaller benefit suggesting they safeguard against poor quality teacher layers, despite preventing a student from fully learning a teacher's knowledge. Notably, there are two cases where standard layer selection with logit losses excluded yields even stronger performance gains than observed from our proposed approach: VGG model pair (top row) on CIFAR10 and CIFAR100. This suggests there is still room to further improve the layer selection algorithm. 

\vspace{-0.5em}\paragraph{Experiment 3: Ablation on Knowledge Quality Metric.}
Finally, we evaluate the importance of each term in our KD metric by choosing layers based on each term independently (i.e., $\mathcal{S}$, $\mathcal{I}$, $\mathcal{E}$) as well as the $\sqrt{\mathcal{I}\mathcal{E}}$ term.  The \emph{only} strategy which \emph{always} successfully converged during training is our method with all three metrics: $\mathcal{Q}$.  The student did not converge in 4 of 12 cases when using only $\mathcal{S}$, $\mathcal{I}$, or $\sqrt{\mathcal{I}\mathcal{E}}$ and in 5 cases when using only $\mathcal{E}$. Notably, teacher models differed in their sensitivity to layer selection. VGG19 proved the most sensitive, with 3 out of the 4 modifications resulting in convergence failures. In contrast, ResNet34 and ViT\_B were more robust, tolerating a broader range of layer selections, with some having a negligible impact on performance.  Even when these layer selection variants successfully converged, they performed worse than our method by $0.5-10\%$.

\section{Limitations}
We only considered image classification in our empirical analyses, as other computer vision tasks such as image segmentation do not have any notion of logits. However, analogous research questions could be explored in the context of both computer vision tasks and classification tasks involving modalities such as text and audio. Due to the computationally intensive nature of our geometric analysis, we did not experiment with larger scale datasets such as Image Net \cite{ImageNet}. Investigating the scaling properties of our proposed method is a valuable direction for future work. Finally, we combined $\mathcal S$, $\mathcal I$, and $\mathcal E$ into the $\mathcal Q$ metric based on empirical evidence. As such, we suspect that the formulation of this metric could be improved with principled theoretical analysis.

\section{Conclusion}
We demonstrated that training the student backbone without any logit-based losses and our geometry-aware novel teacher layer selection method achieves state-of-the-art KD performance. This challenges the well-established paradigm of augmenting feature-based losses with logit-based losses. Directions for future work include: (1) exploring how the $\mathcal{S}$, $\mathcal{I}$, and $\mathcal{E}$ measures relate to model generalization outside of KD, (2) studying how student layer selection influences distillation, and (3) investigating if other tasks benefit from similar loss recipe refinements.

\paragraph{Acknowledgments.}
We thank Fran\c{c}ois Meyer and James Meiss for insightful discussions. 

\small
\bibliographystyle{plainnat}
\bibliography{main.bib}

\clearpage
\appendix
\section{Theoretical Appendix}

\subsection{Hyperspherical Packing}
We provide here the proof of the result on hyperspherical packing that is used to define efficiency $\mathcal{E}$. Recall the ReLU activation function:
\begin{equation*}
    ReLU(x) = \max\{x, 0\}.
\end{equation*}
The $D$-dimensional ReLU activated space is thus the non-negative orthant $\mathbb{R}^D_{+}$. 

\paragraph{Theorem:} Given a set of $N$ distinct points $p_i$ in $D$-dimensional ReLU activated space:
\begin{equation*}
    P = \left\{ p_1,~p_2, ~p_3, \cdots,~p_N\right\} 
\end{equation*}
and a minimum distance $d_{min} > 0$ between these points, the smallest radius of the $D$-dimensional hypersphere $S^{D-1}$ centered at the origin that can accommodate the set $P$ of points can be approximated by:

\begin{equation*}
    r_{min} \approx 2d_{min} \left( \frac{N}{\pi} \right)^{\frac{1}{D-1}}. 
\end{equation*}

\noindent
\textbf{Proof:} Note that the surface area of hypersphere $S^{D-1}$ of radius $r$ is given by: 
\begin{equation*}
    Surf(S^{D-1}(r)) = r^{D-1}\frac{2\pi^{\frac{D+1}{2}}}{\Gamma(\frac{D+1}{2})}
\end{equation*}
where $\Gamma(\cdot)$ is the gamma function. We approximate the problem by finding how many $(D-1)$-balls of radius $d_{min}$ fit in the surface area. This is a good approximation when the number of points $N$ is large; e.g., $N \geq 50,000$ for CIFAR10/100 and Tiny ImageNet. For a proper treatment of the asymptotic behavior (with respect to $N$) of this question, see \cite{HypersphereDetails}. The volume of a $(D-1)$-ball is given by:
\begin{equation*}
    Vol(B^{D-1}(d_{min})) = d_{min}^{D-1}\frac{\pi^{\frac{D-1}{2}}}{\Gamma(\frac{D+1}{2})}.
\end{equation*}
The positive orthant only accounts for a factor of ${2^{-D}}$ of the hypersphere. 
So, the number of $(D-1)$-balls of radius $d_{min}$ that can fit on the non-negative orthant portion of the hypersphere of radius $r$ is approximated by:
\begin{equation*}
\begin{aligned}
    N &\approx \frac{2^{-D} Surf(S^{D-1}(r))}{Vol(B^{D-1}(d_{min}))}\\
    &= 2^{-D} \frac{r^{D-1}}{d_{min}^{D-1}}\frac{2\pi^{\frac{D+1}{2}}}{\pi^{\frac{D-1}{2}}}\\
    &= \pi \left(\frac{r}{2d_{min}}\right)^{D-1}, 
\end{aligned}
\label{eq:Surf/Vol}
\end{equation*}
from which we obtain: 
\begin{equation*}
    r \approx 2d_{min}\left(\frac{N}{\pi}\right)^{\frac{1}{D-1}}.
\end{equation*}
 
\paragraph{Remark:} We consider ReLU because of its popularity and simple geometric interpretation and analytic properties.  Similar results can be obtained for other activation functions, such as Leaky ReLU and GeLU. 

\subsection{Intrinsic and Embedding Dimension Background}
\begin{figure}[h!]
	\centering
	\includegraphics[scale = 0.12]{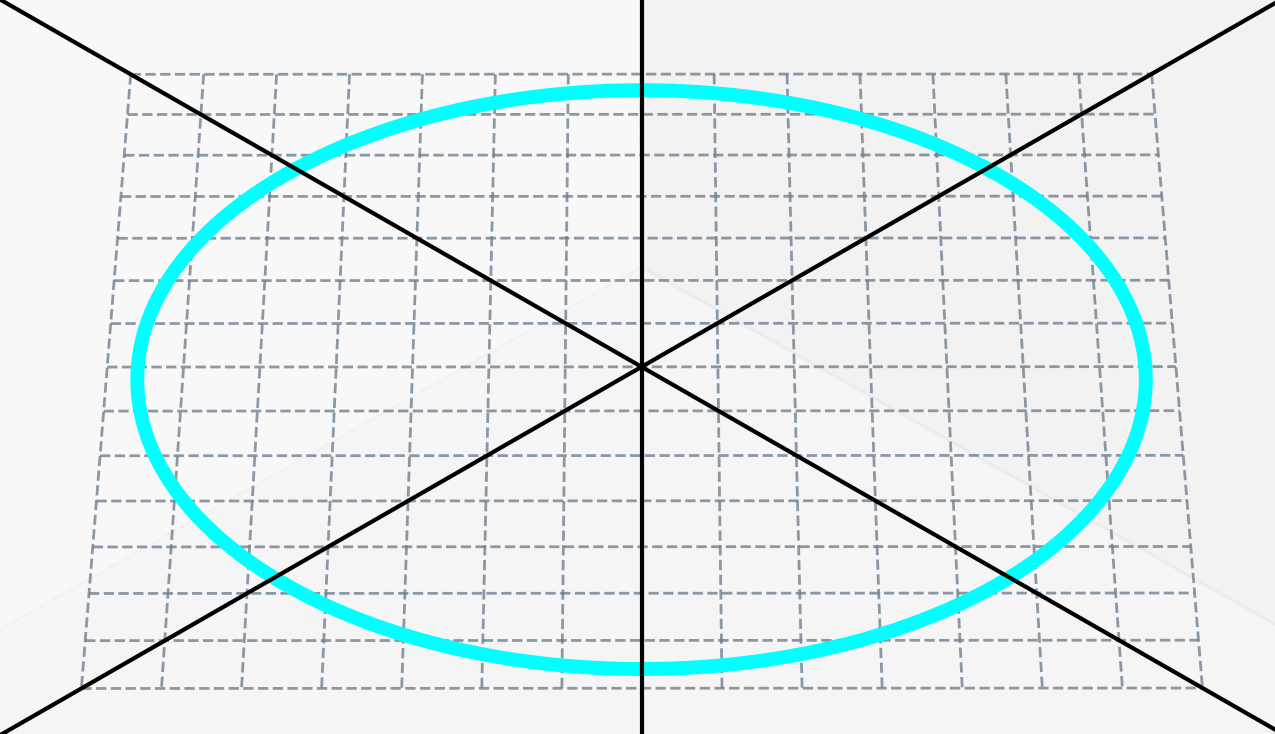}
	\caption{Relationship between the intrinsic, embedding, and ambient dimensions (ID, ED, AD). The blue circle has ID $1$, because it is a 1-dimensional manifold. However, it has ED 2 (dashed lines), because it cannot exist in $R^n$, when $n < 2$. Yet, the circle is drawn in AD 3. Generally, $ID \leq ED \leq AD$.}
    \label{fig:ED-Example}
\end{figure}

To facilitate understanding of the important geometric concept of intrinsic dimension, we provide a visualization of the differences between the ambient, intrinsic, and embedding dimensions in \textbf{Figure \ref{fig:ED-Example}}. Pictured there is 3-D Euclidean space with two embedded sub-manifolds; the circle and the plane. Because the circle has a ``true'' dimension of 1, its ID is 1. Intuitively, this is because at any point on the circle, there are only two possible directions to move; counter-clockwise and clockwise. However, because we cannot draw circles in 1-D Euclidean space, the embedding dimension of the circle is 2.

\begin{figure}[h!]
    \centering
    \includegraphics[scale = 0.079]{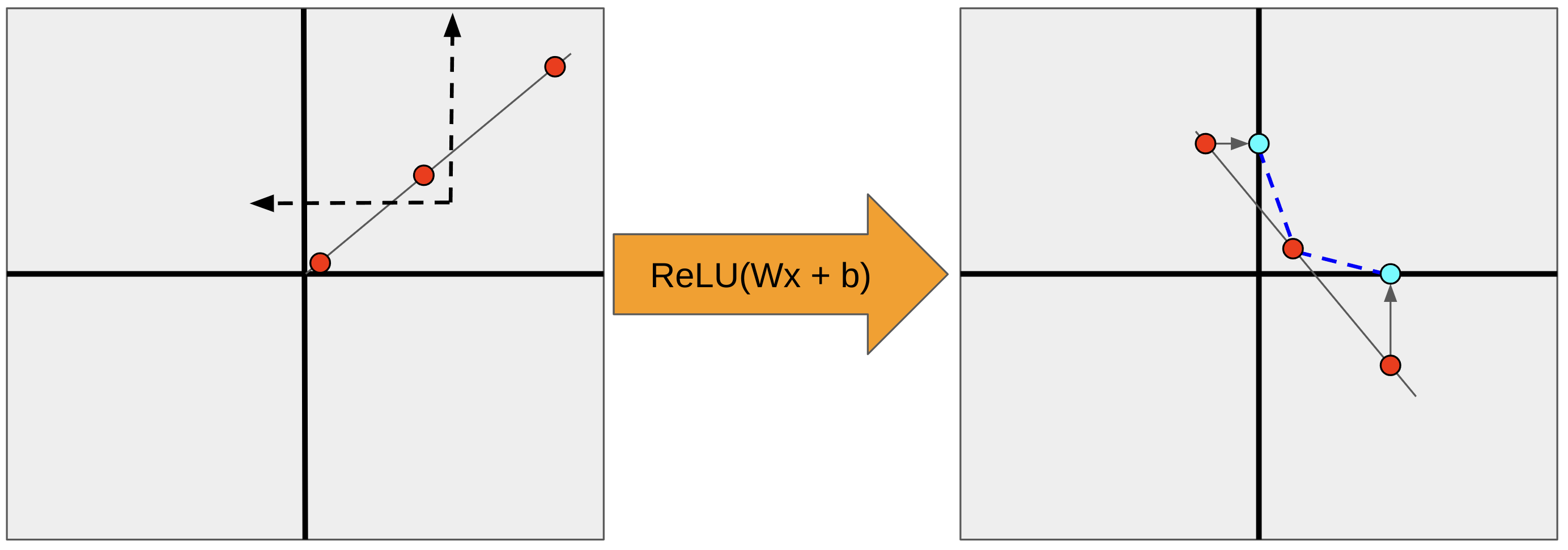}
    \caption{Example of increased embedding dimension as a result of ReLU activation.}
    \label{fig:EDEx}
\end{figure}

Visualizing what an increase in embedding dimension might look like can be non-obvious at first glance. To see how embedding dimension can increase via processing in non-linear neural networks, consider \textbf{Figure \ref{fig:EDEx}}.

\noindent
Starting on the left, $3$ points on a line in $\mathbb{R}^2$ (red dots) with embedding dimension $1$. Then, a new basis is defined by a linear layer (dashed lines) and the red dots are reinterpreted in this new basis (right). ReLU activation is applied (blue dots), increasing the embedding dimension to $2$ (dashed blue lines). Intuitively, the data are ``bent" around a mold of shape defined by the activation function. The position of this mold is determined by the learned change of basis.  
\section{Technical Appendix}

\subsection{Model Architecture Details}
\paragraph{ResNet9.} ResNet18 consists of $2$ residual blocks \cite{ResNet} per stage, meaning there are $2$ blocks for each spatial dimension of the representations. ResNet9 is formed simply by reducing this to $1$ residual block per stage. All other layers, such as the initial convolution prior to the residual blocks, remain unaltered. 

\paragraph{ViT\_ET.} For the ``extra tiny" vision transformer, we follow the structure of ViT\_B \cite{ViT} but reduce the number of layers, hidden dimension, and number of attention heads. Recall that ViT\_B has $12$ layers with $12$ attention heads, and a hidden dimension of $768$. ViT\_ET has $8$ layers with $6$ attention heads, and a hidden dimension of $192$. This results in a total of $\sim 3M$ parameters, which is comparable to MobileNetV2 \cite{MobileNet} at $2.5M$ parameters. For additional context, VGG11 \cite{VGG} has $28M$ and ResNet9 has $5M$.  

\begin{table}
    \centering
    \begin{tabular}{c|c}
        Model & MaxLR \\
        \hline
        VGG19 & 0.005\\
        VGG11 & 0.005\\
        \hline
        ResNet34 & 0.0075\\
        ResNet9 & 0.0075\\
        MobileNetV2 & 0.0075\\
        \hline
        ViT\_B & 0.0001\\
        ViT\_ET & 0.001
    \end{tabular}
    \caption{Maximum learning rates for each architecture. Kept fixed for all datasets and all KD configurations. }
    \label{tab:MaxLR}
\end{table}

\begin{figure*}[t!]
	\includegraphics[width=\textwidth, height = 8cm]{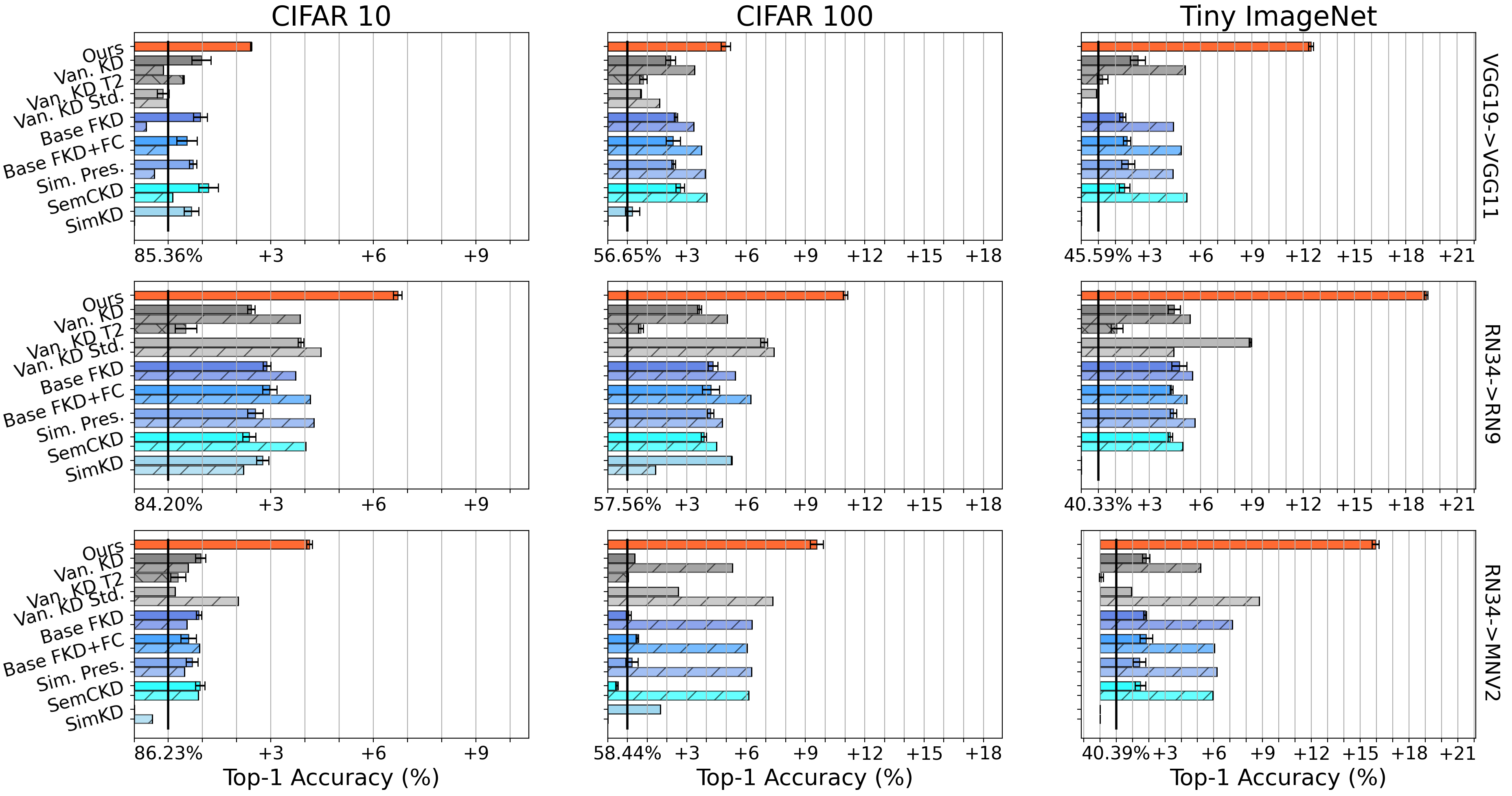}
	\caption{Training protocol validation. Our method is trained for $50$ epochs. Baselines are given $\sim5\times$ training time of $240$ epochs (striped bars). VKD-T2 denotes Vanilla KD w/ $T=2$. Configurations which failed to converge are not plotted.}
	\label{Experiment1V}
\end{figure*}

\begin{figure*}
	\includegraphics[width=\textwidth, height = 8cm]{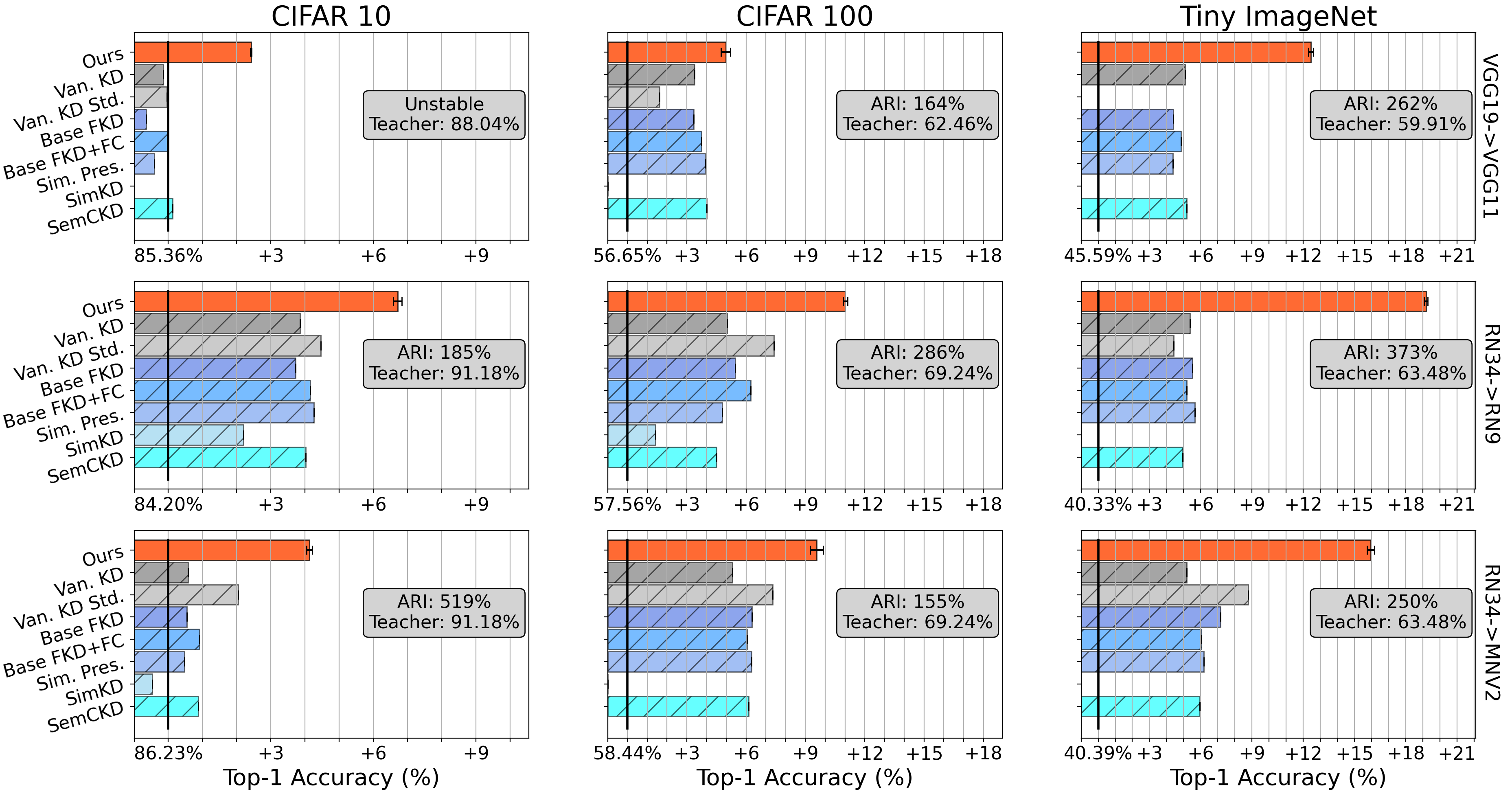}
	\caption{Training protocol validation ARIs. Our method is trained for $50$ epochs. Baselines are given $\sim5\times$ training time of $240$ epochs (striped bars). Unstable ARI indicates unrepresentative values due to poor baseline performance.  Configurations which failed to converge are not plotted.}
	\label{Experiment1V2}
\end{figure*}

\subsection{Optimization Details \label{OptimDetails}}
We used the Adam \cite{Adam} optimizer to train all models. Weight decay was set to $0.01$, $\beta_1 = 0.9$, $\beta_2 = 0.999$, and the numerical stabilization term was set to $1e-8$. We trained all models for $50$ epochs.  We used the One Cycle learning rate scheduler \cite{OneCycle}, with its suggested technique for choosing the maximum learning rate; i.e., training baseline models (without any distillation) with a range of different maximum learning rates for $25$ epochs to select the optimal value. This was done for each architecture, and then the result was fixed for all experiments. See \textbf{Table \ref{tab:MaxLR}} for the specific values. 

The other parameters of the scheduler are as follows: $30\%$ of the iterations were spent increasing the learning rate ($\eta$), at the start of training $\eta = \frac{MaxLR}{25}$, and at the end $\eta = \frac{MaxLR}{10000}$. We used cosine annealing, with base and peak momentums set to $0.85$ and $0.95$, respectively.

\subsection{Extra Validation Experiments}

\paragraph{Is 50 Epochs Long Enough?}
We now justify our training approach (\textbf{Section \ref{OptimDetails}}) by showing our findings from experiment 1 of the main paper are consistent even when leveraging other training strategies.  Specifically, we follow prior work~\cite{TeacherAttnKD, InfMatchKD, ReusedClsKD} by also training student models for $240$ epochs using stochastic gradient descent (SGD). This analysis is intended to address potential concerns that our findings are due to baseline methods not properly converging.  We call this the ``standard" optimization scheme. We use momentum of $0.9$ following a multi-step learning rate scheduler, where the learning rate was stepped down by a factor of $10$ at epochs $150$, $180$, and $210$. For ResNet9 and MobileNetV2, the initial learning rate was $0.05$, while for VGG11 it was set to $0.01$ because $0.05$ resulted in failure to converge. We omit the vision transformer model pair (ViT\_B $\rightarrow$ ViT\_ET) from this experiment because we observed significant performance degradation ($\sim -10\%$) in the student model when using this optimization scheme. 

Results are shown in \textbf{Figure \ref{Experiment1V}}, with updated ARI values provided in \textbf{Figure \ref{Experiment1V2}}. Despite being provided with almost $5\times$ more training time, none of the baseline techniques surpass the performance of our method presented in the main paper.  Instead, we still observe the same conclusions as articulated in the main paper: (1) our method achieves superior performance, (2) its best results are on Tiny ImageNet, and (3) its worst results are on the vision transformer model pair. See \textbf{Section \ref{TrEff}} training curves of the student models.  Additionally, our method often surpasses the (non-distillation) student's accuracy after only $10-30$ epochs. This is distinctly different behavior from all baseline methods, which appear to demand significantly increased computational budget. 

\paragraph{Is $T=4$ Too High?}
We also investigated if an overly high temperature could be the cause of poor baseline performance. This is inspired in part by our observation during baseline training that the $\mathcal{L}_{KL}$ loss term was sometimes larger than the cross-entropy loss. We re-trained VKD with a reduced temperature of $2$. This configuration is denoted by ``VKD-T2". Results are shown in \textbf{Figure \ref{Experiment1V}}.  We observe that the temperature choice of $T=4$ is superior to $T=2$, with the reduced temperature resulting in considerable performance degradation in most cells. 

\begin{figure*}
	\includegraphics[width=\textwidth, height = 8cm]{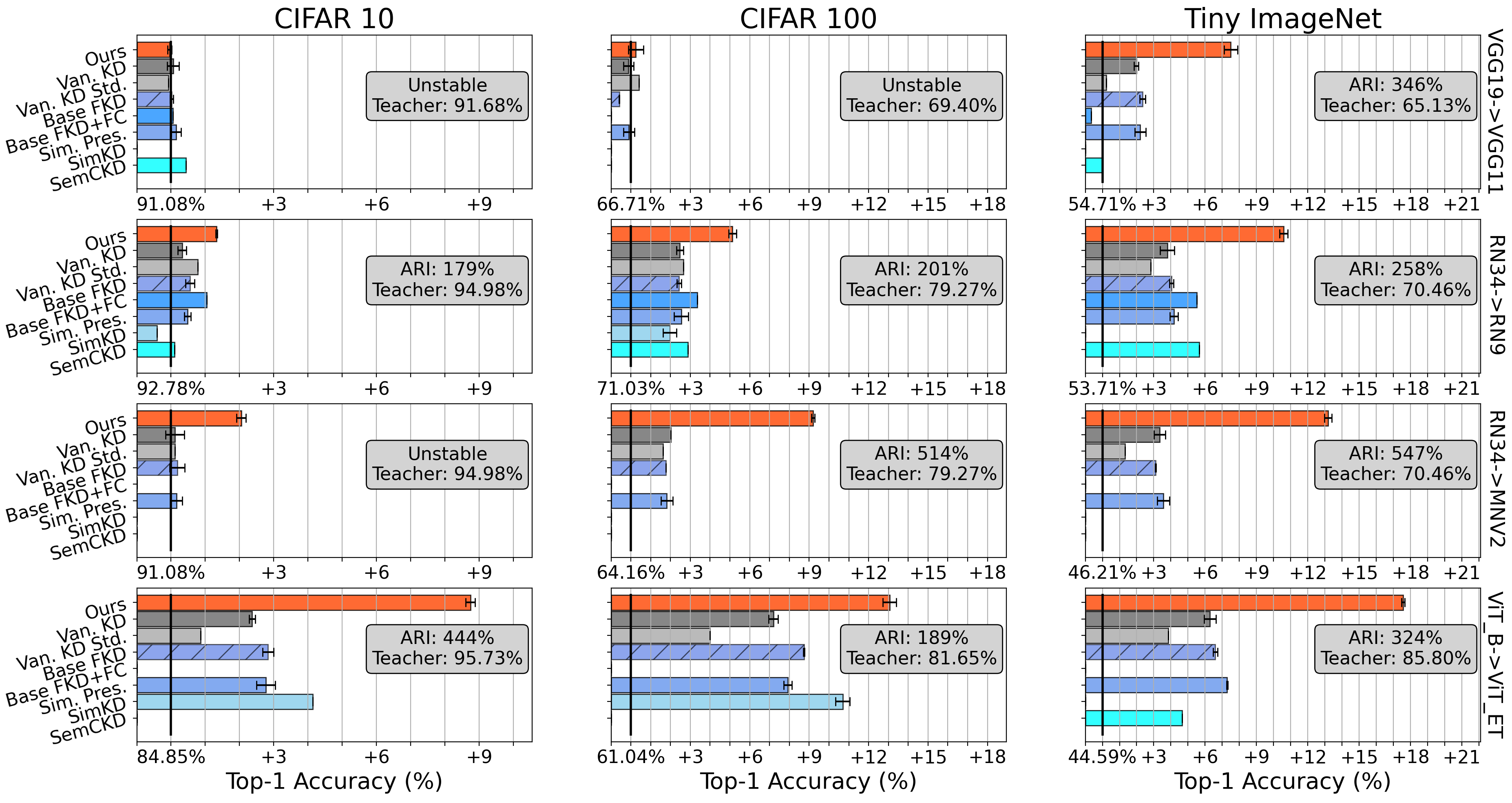}
	\caption{Data Augmentation validation results. Configurations which failed to converge are not plotted. Unstable ARI indicates unrepresentative values due to proximity to baseline student.}
	\label{Experiment1DA}
\end{figure*}

\paragraph{Data Augmentation Experiments.}
As discussed in the main paper, we did not use data augmentation in the main experiments to ensure reproducible knowledge quality computation. To validate our method's effectiveness when trained with data augmentation, we re-trained all configurations with a standard data augmentation recipe of random horizontal flips (with $50\%$ probability) and random crops from zero padded images. We used a padding size of $4$ on all sides of the images. \textbf{Figure \ref{Experiment1DA}} contains the results.

Once again, we re-observe the same conclusions articulated in the main paper. That is, our method outperforms all the baselines, and achieves its best (relative) results on Tiny Imagenet and worst results on the ViT model pair. In fact, our method achieves a new best recorded result for MobileNetV2 on CIFAR100 at $73.36\%$ accuracy.

\section{Complete Results}
\subsection{Training Curves \label{TrEff}}
\begin{figure}[h]
    \centering
    \includegraphics[scale=0.08]{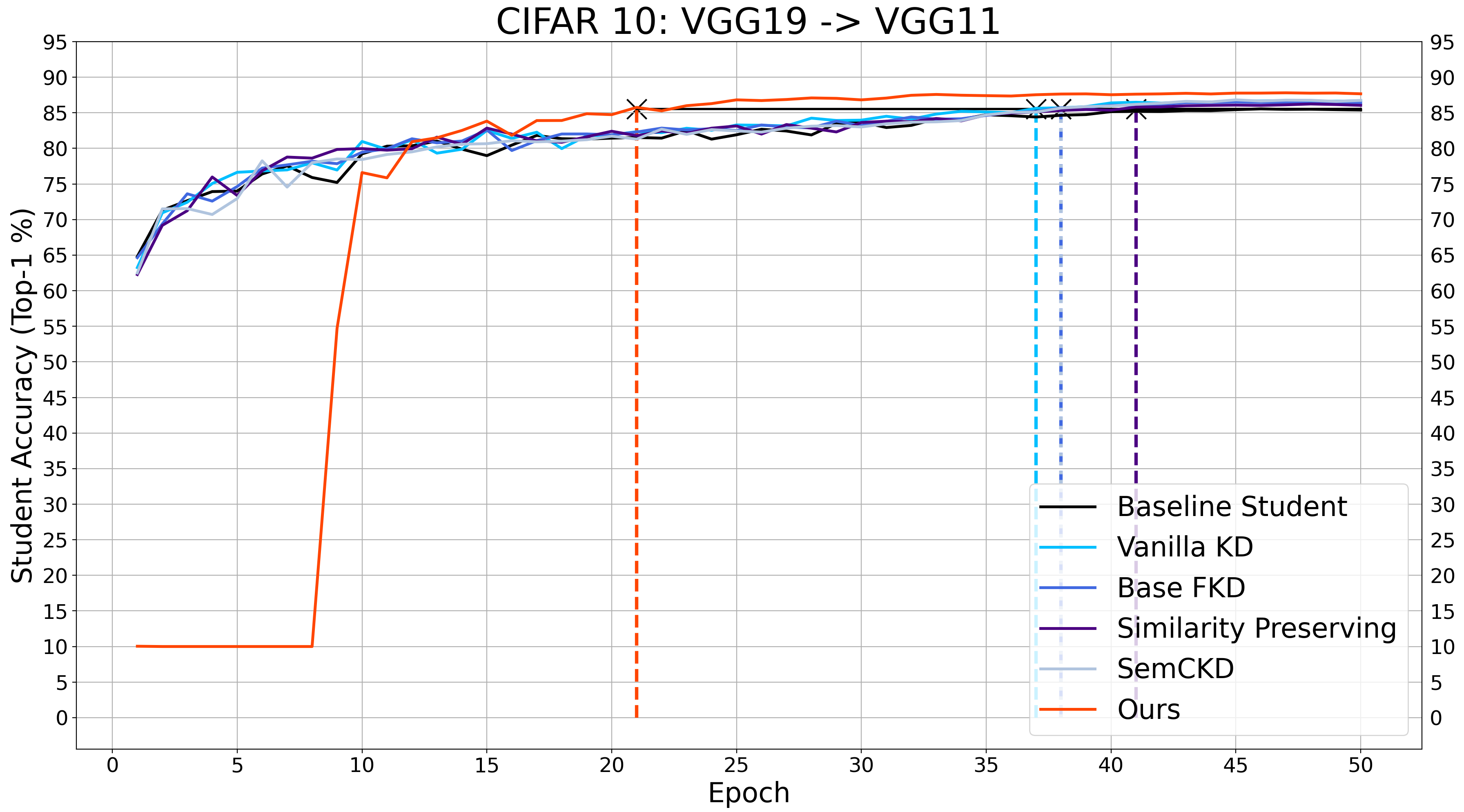}\\
    \includegraphics[scale=0.08]{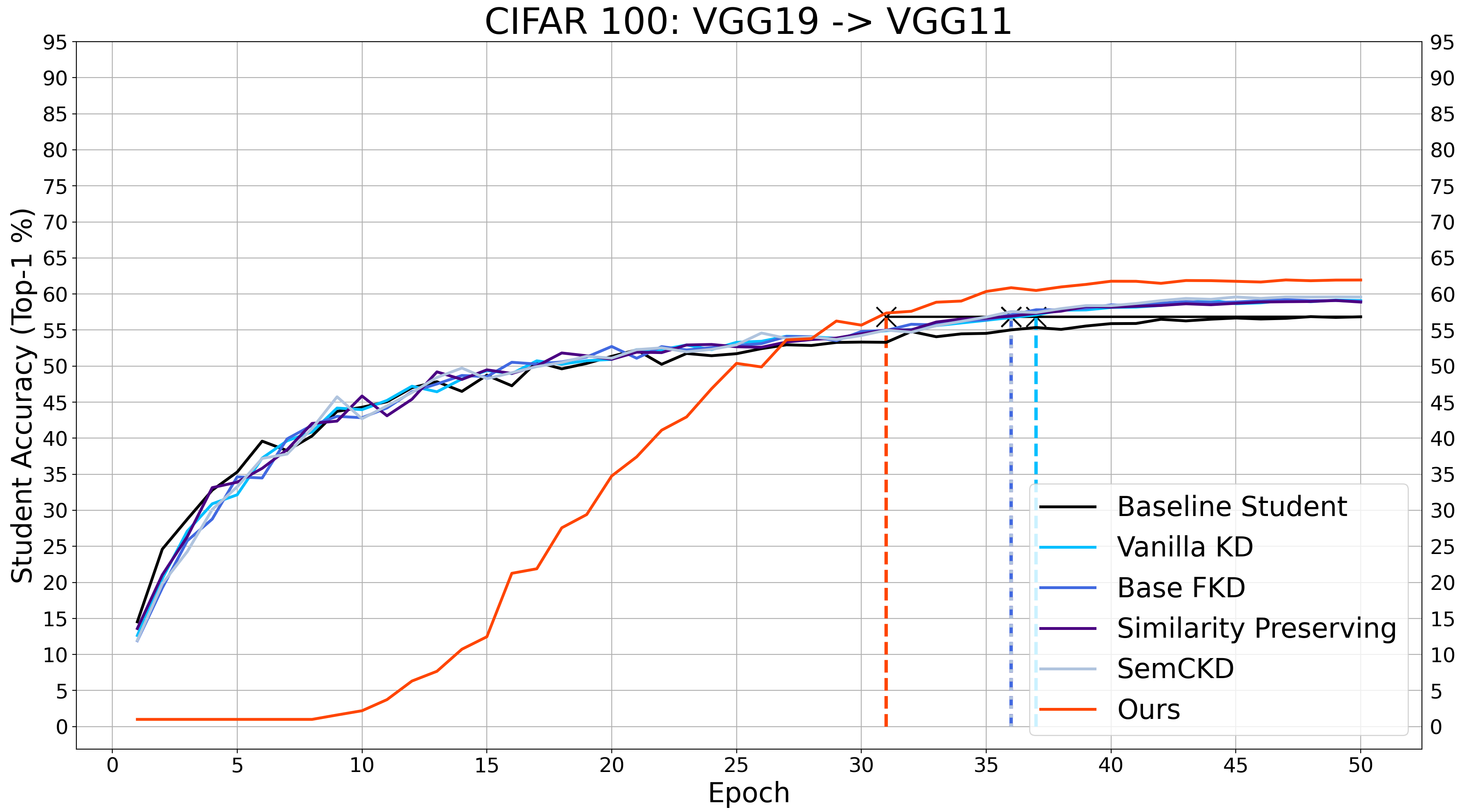}\\
    \includegraphics[scale=0.08]{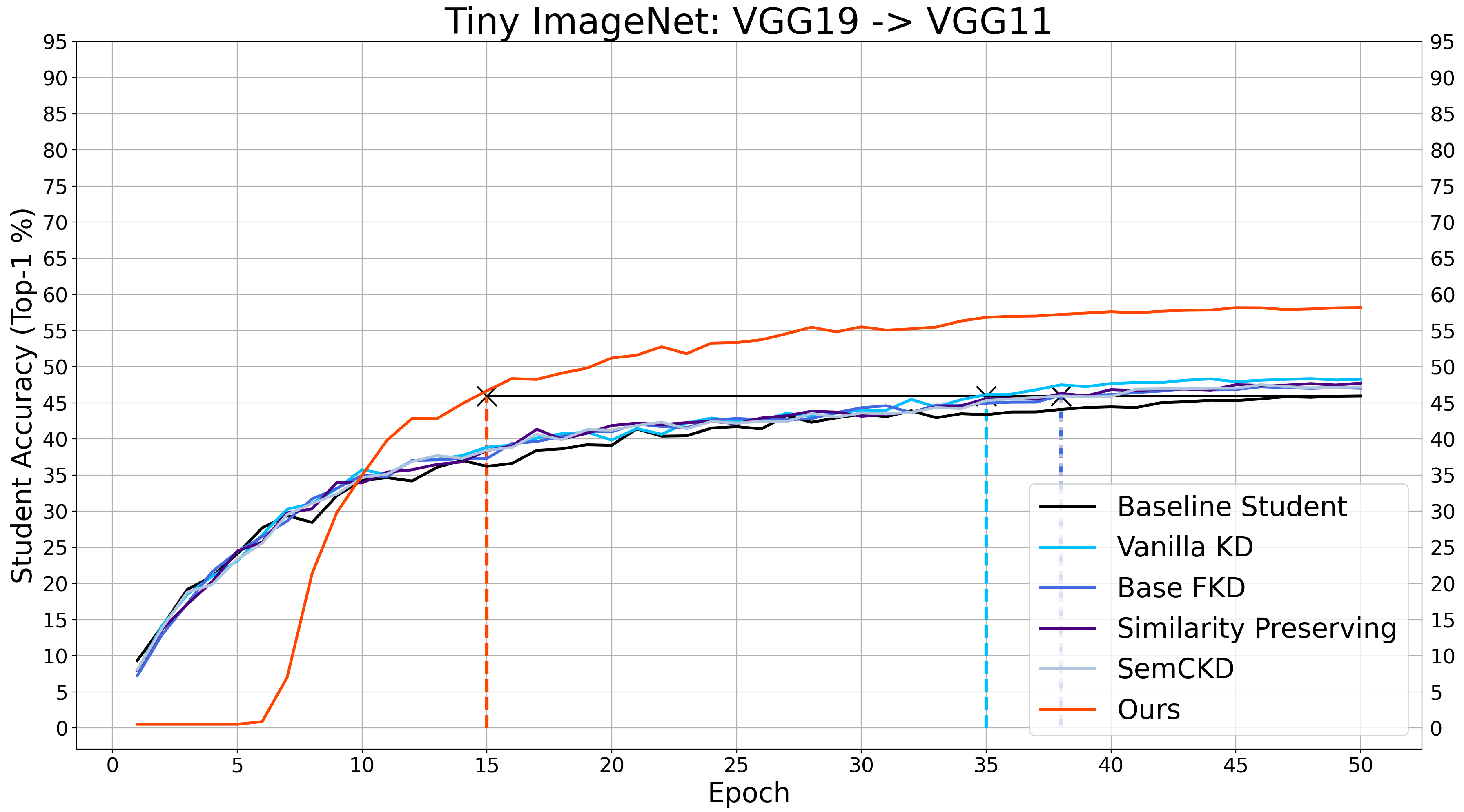}\\
    \caption{VGG19 $\rightarrow$ VGG11 Student Training Curves. X-axis: epochs. Y-axis: top-1 accuracy. Our method shown in orange. Vertical lines indicate when each distillation method surpasses the baseline student's accuracy.}
    \label{fig:VGG11Curves}
\end{figure}

\begin{figure}
    \centering
    \includegraphics[scale=0.08]{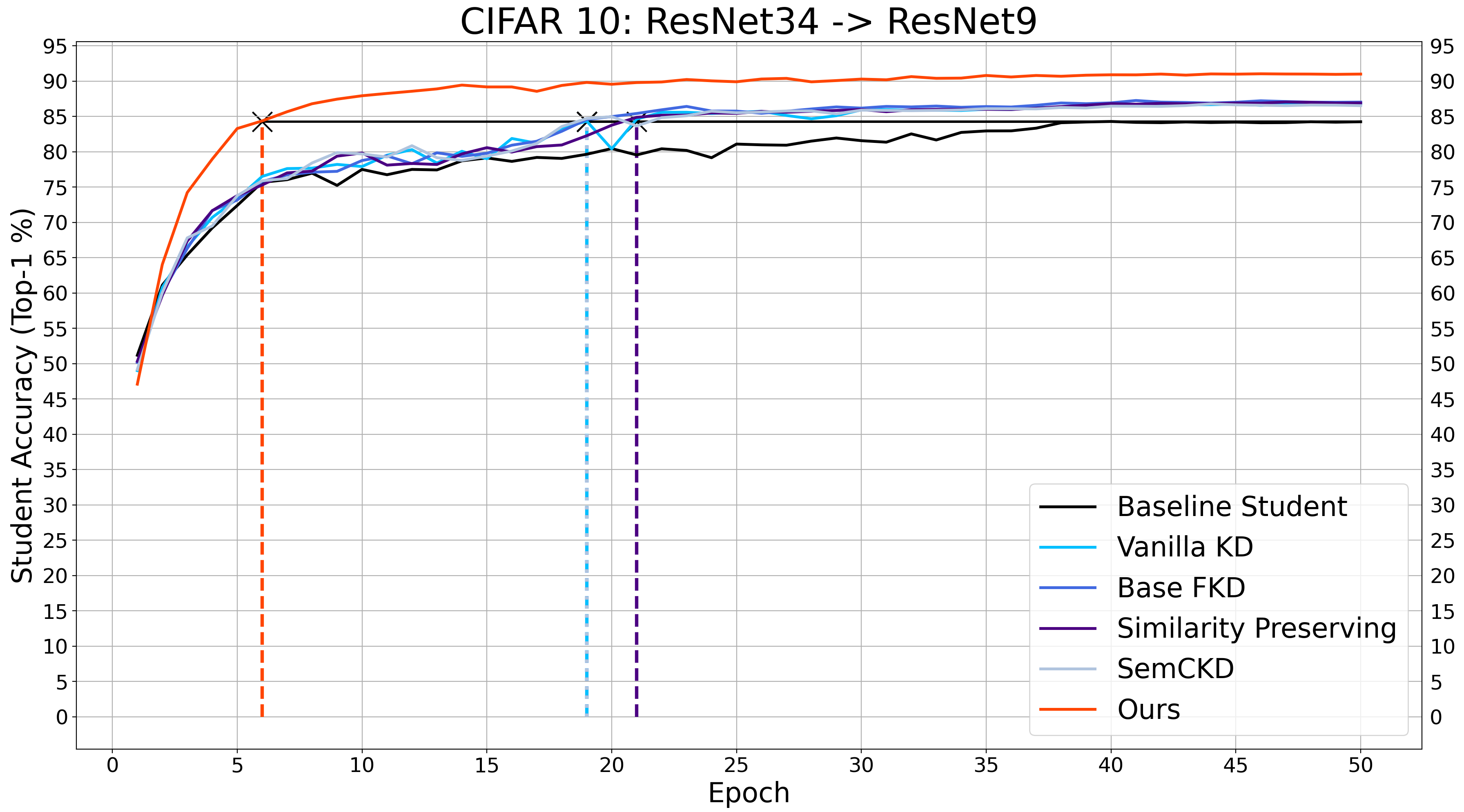}\\
    \includegraphics[scale=0.08]{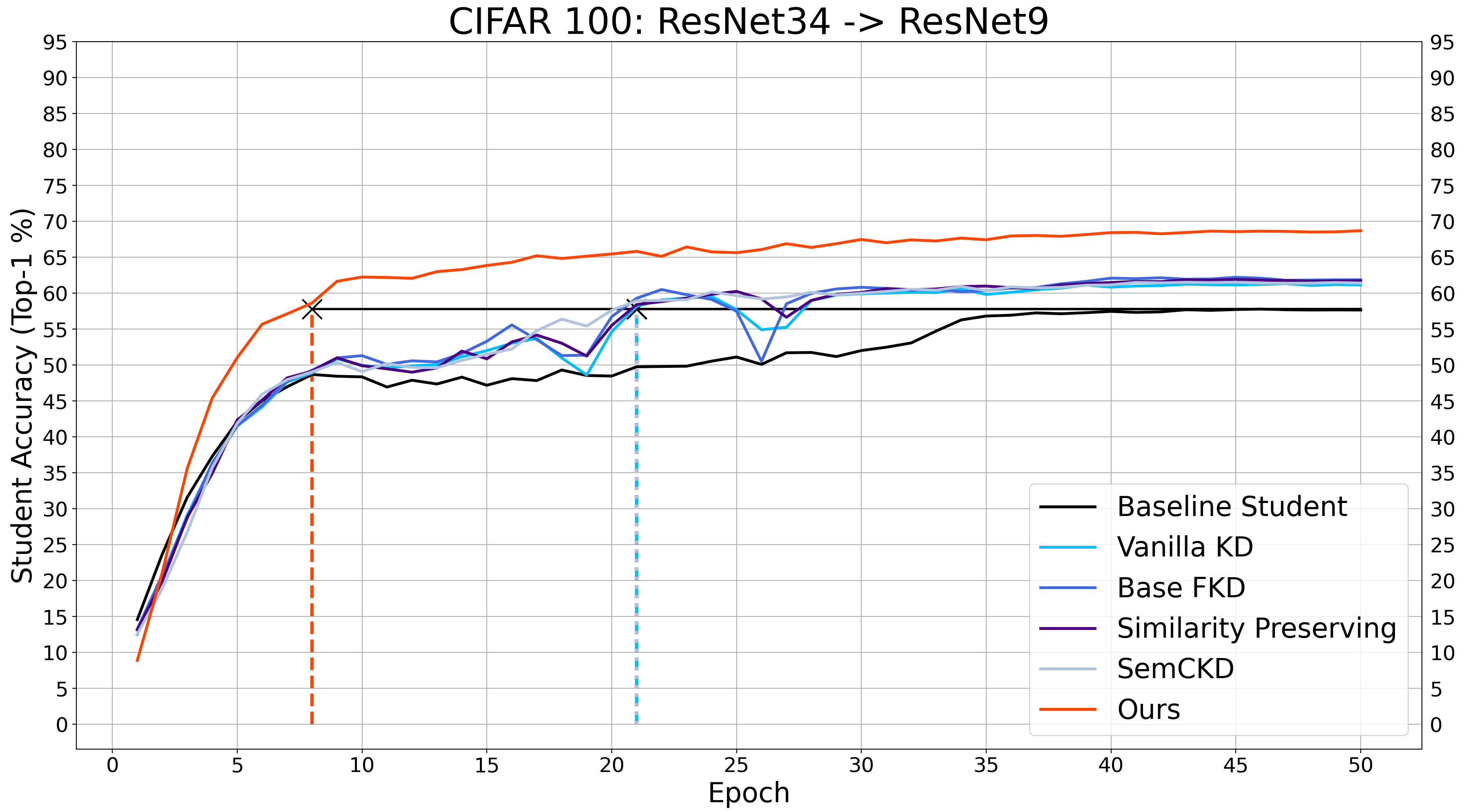}\\
    \includegraphics[scale=0.08]{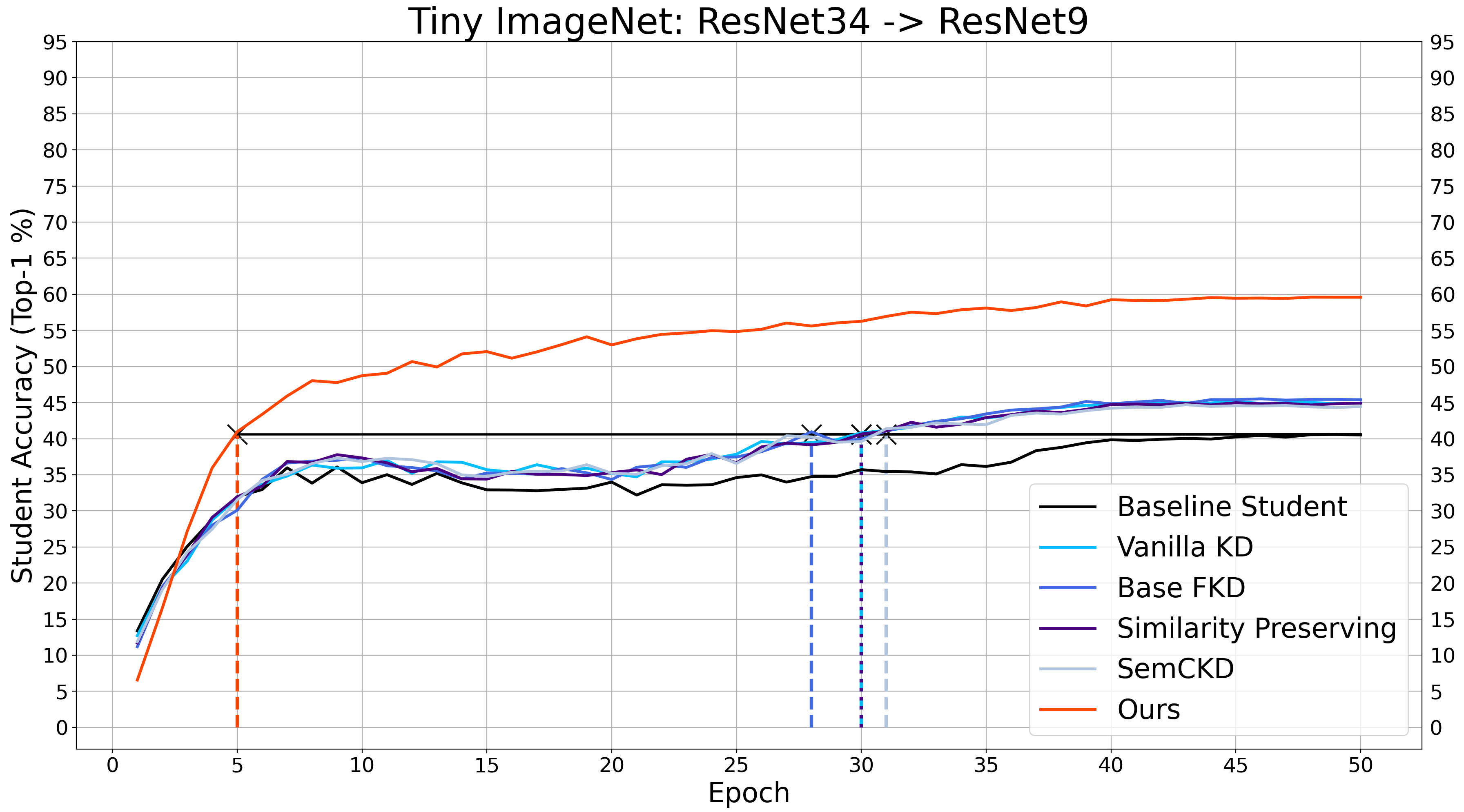}\\
    \caption{ResNet34 $\rightarrow$ ResNet9 Training Curves. X-axis: epochs. Y-axis: top-1 accuracy. Our method shown in orange. Vertical lines indicate when each distillation method surpasses the baseline student's accuracy.}
    \label{fig:ResNet9Curves}
\end{figure}

\begin{figure}
    \centering
    \includegraphics[scale=0.08]{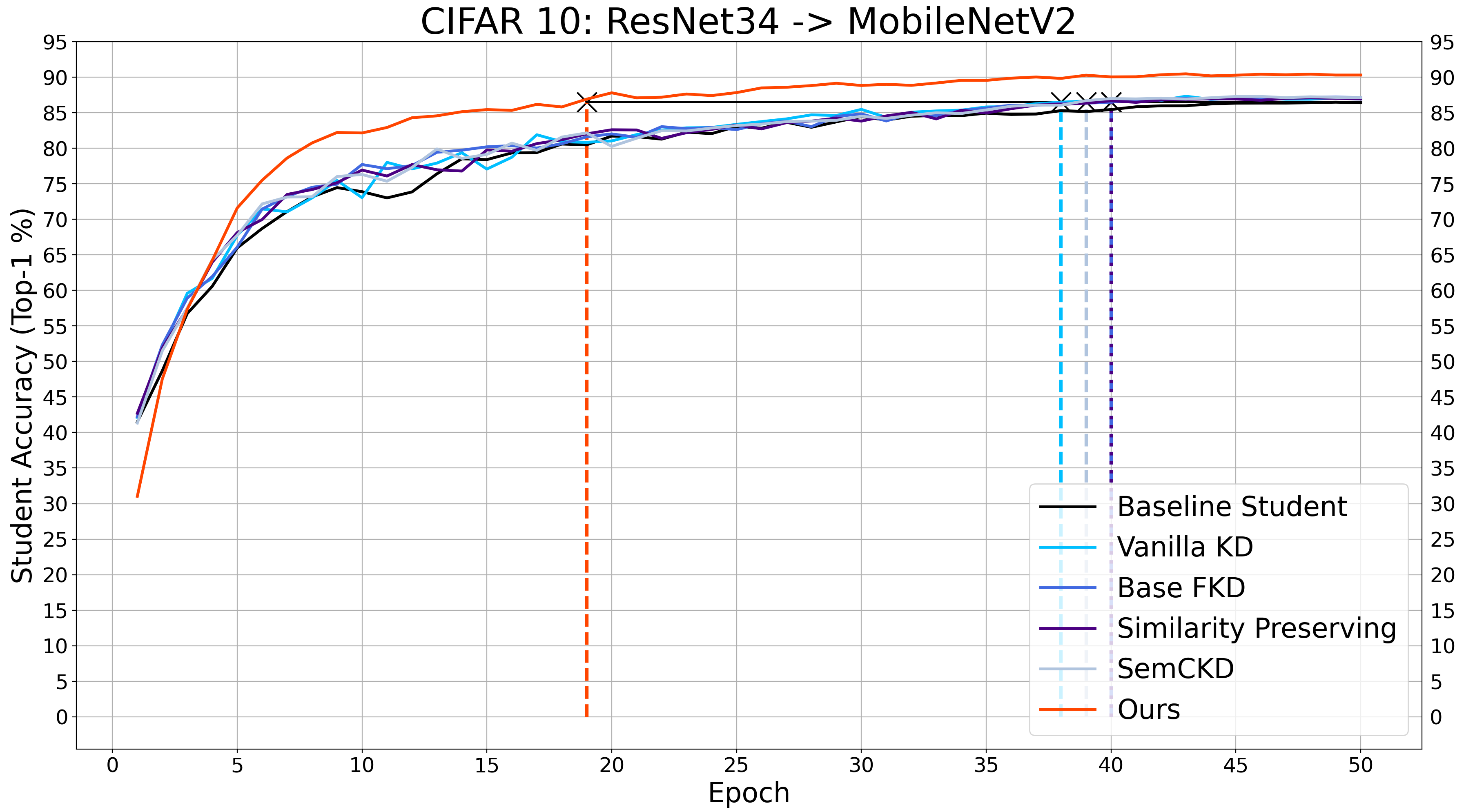}\\
    \includegraphics[scale=0.08]{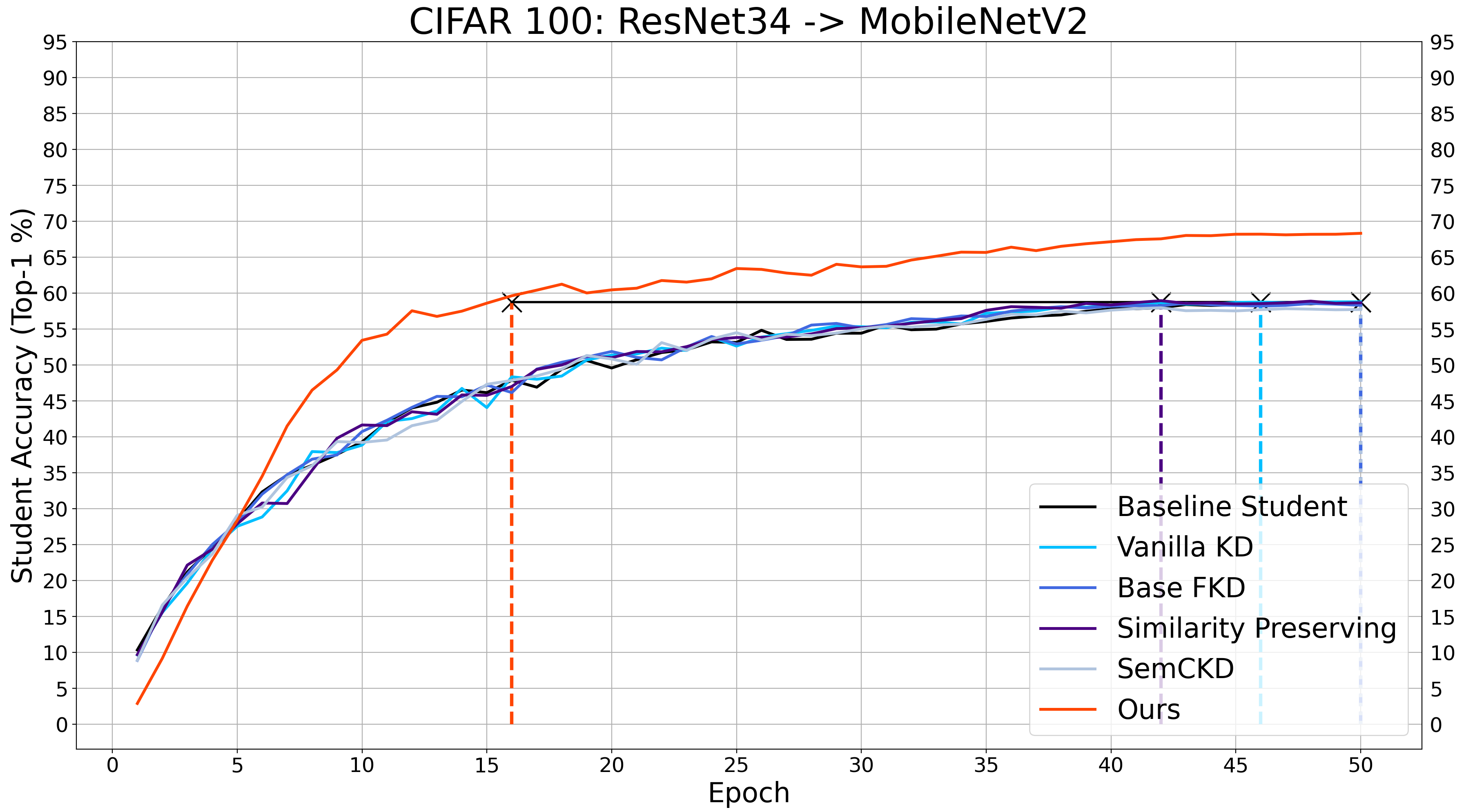}\\
    \includegraphics[scale=0.08]{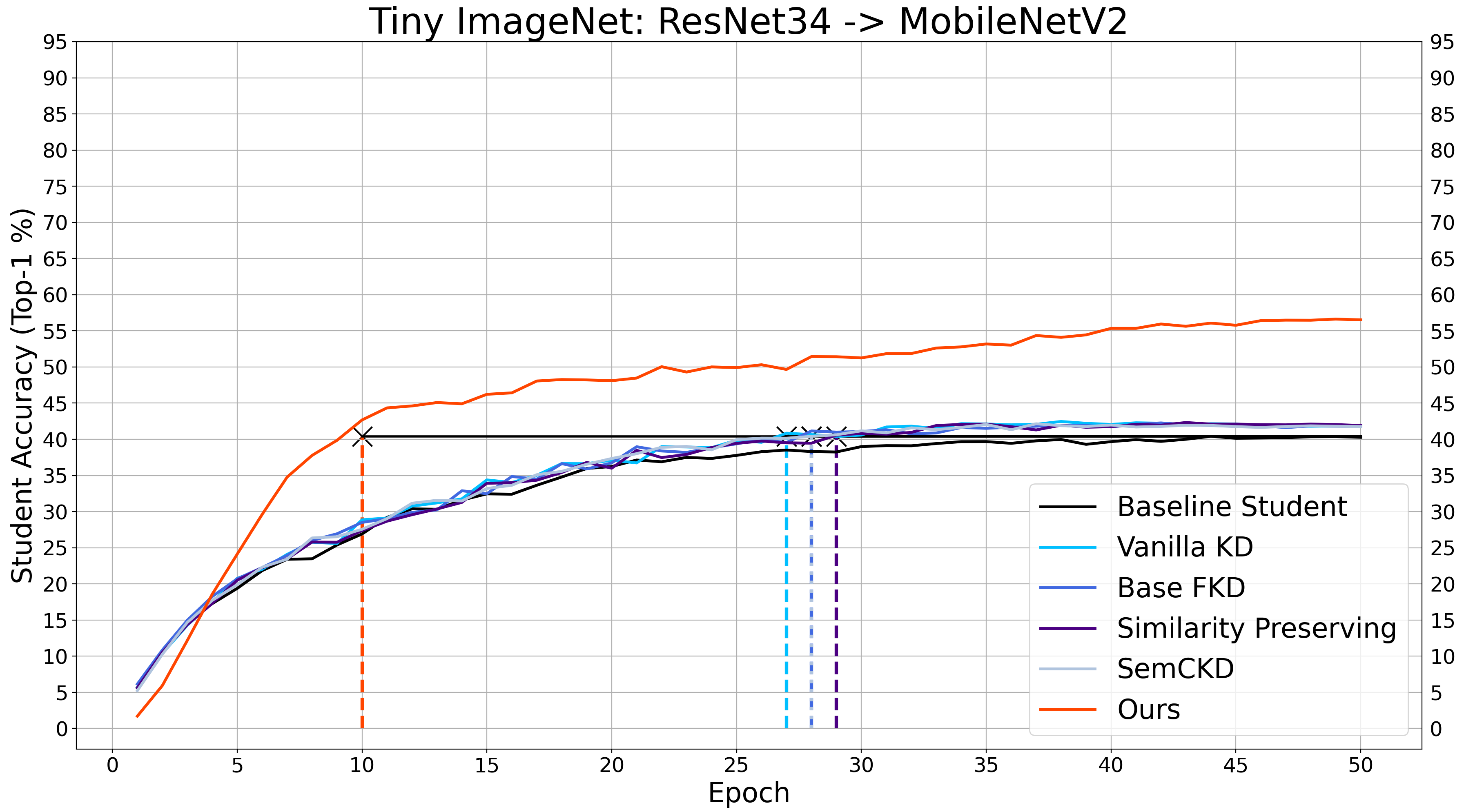}\\
    \caption{ResNet34 $\rightarrow$ MobileNetV2 Training Curves. X-axis: epochs. Y-axis: top-1 accuracy. Our method shown in orange. Vertical lines indicate when each distillation method surpasses the baseline student's accuracy.}
    \label{fig:MN2Curves}
\end{figure}

\begin{figure}
    \centering
    \includegraphics[scale=0.08]{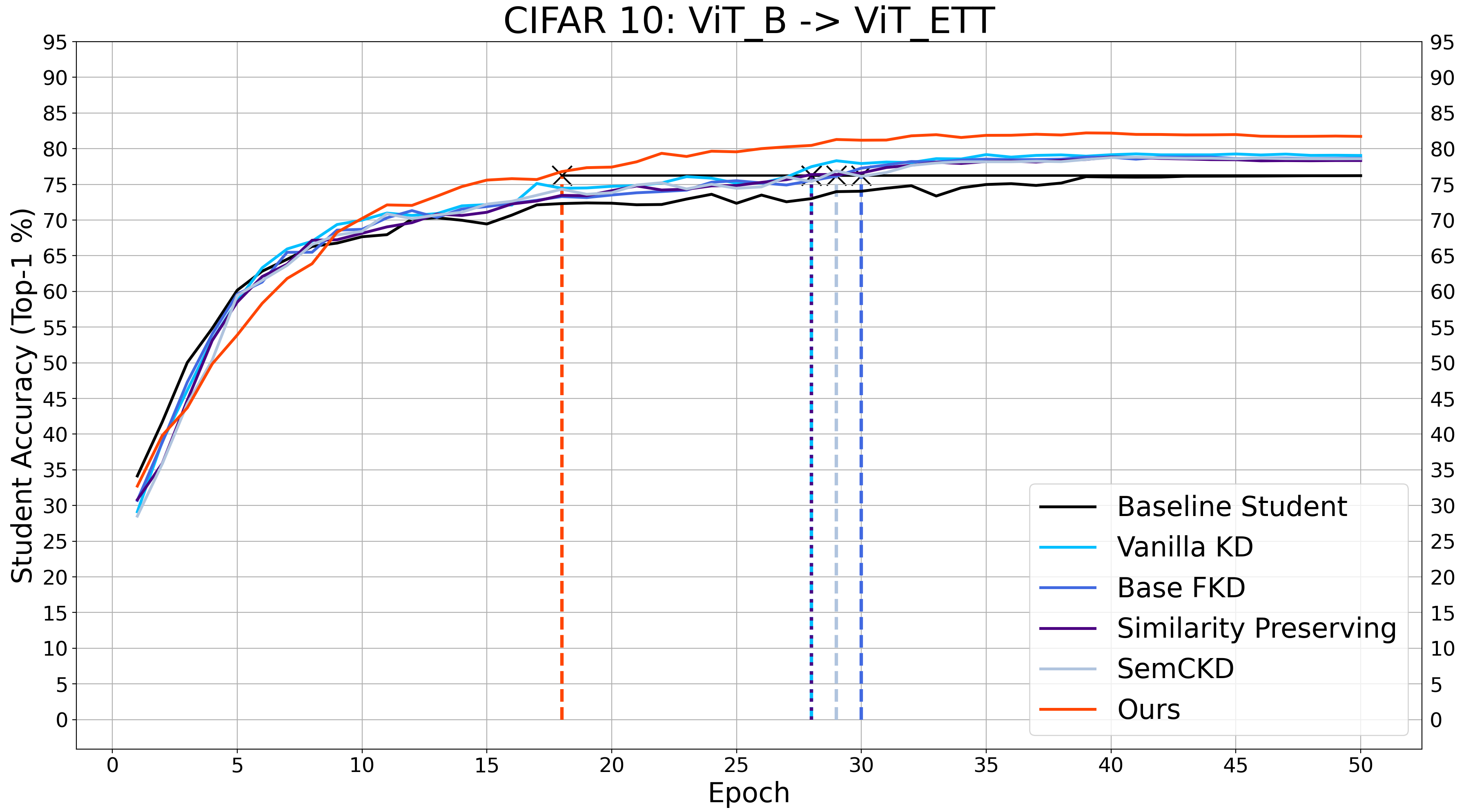}\\
    \includegraphics[scale=0.08]{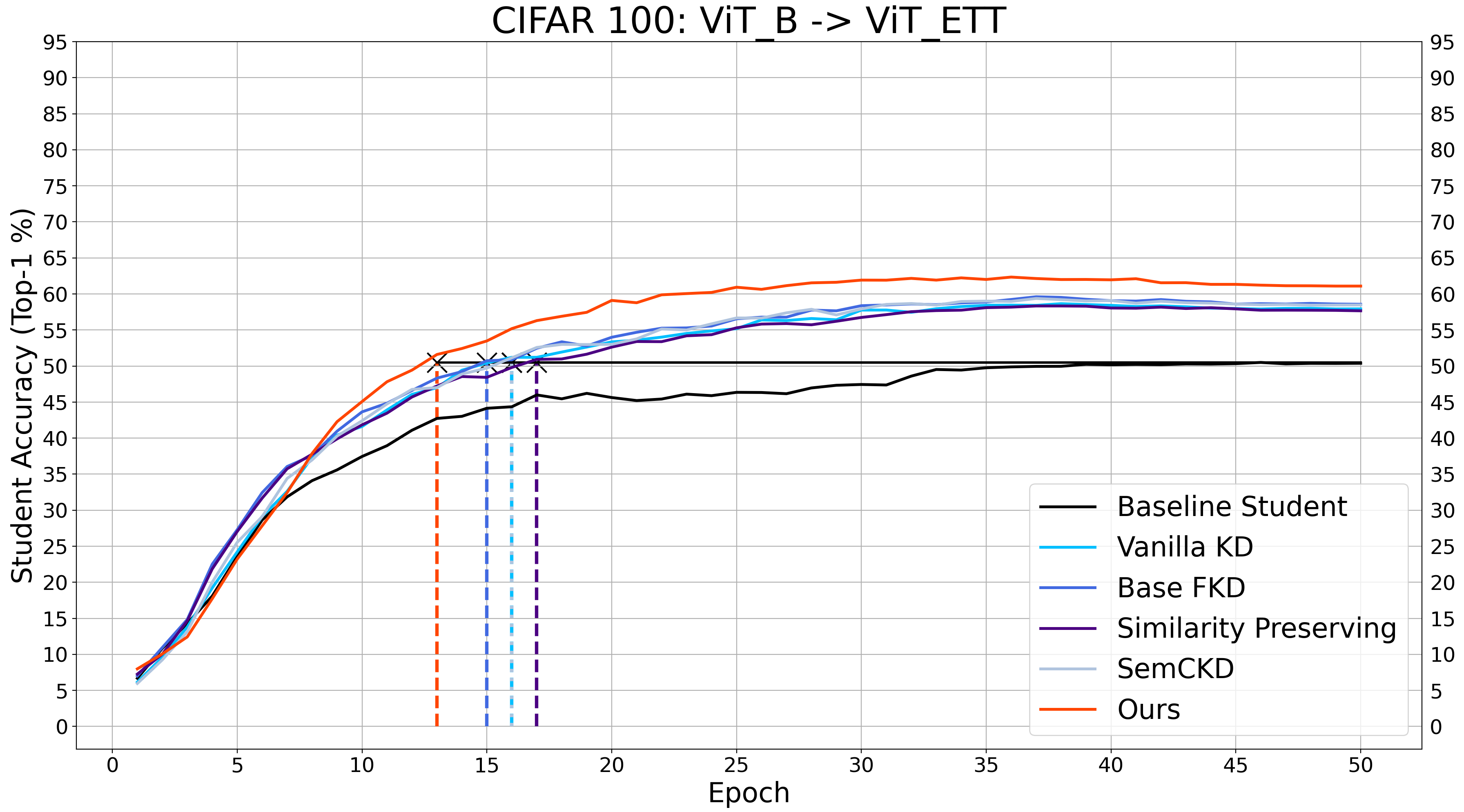}\\
    \includegraphics[scale=0.08]{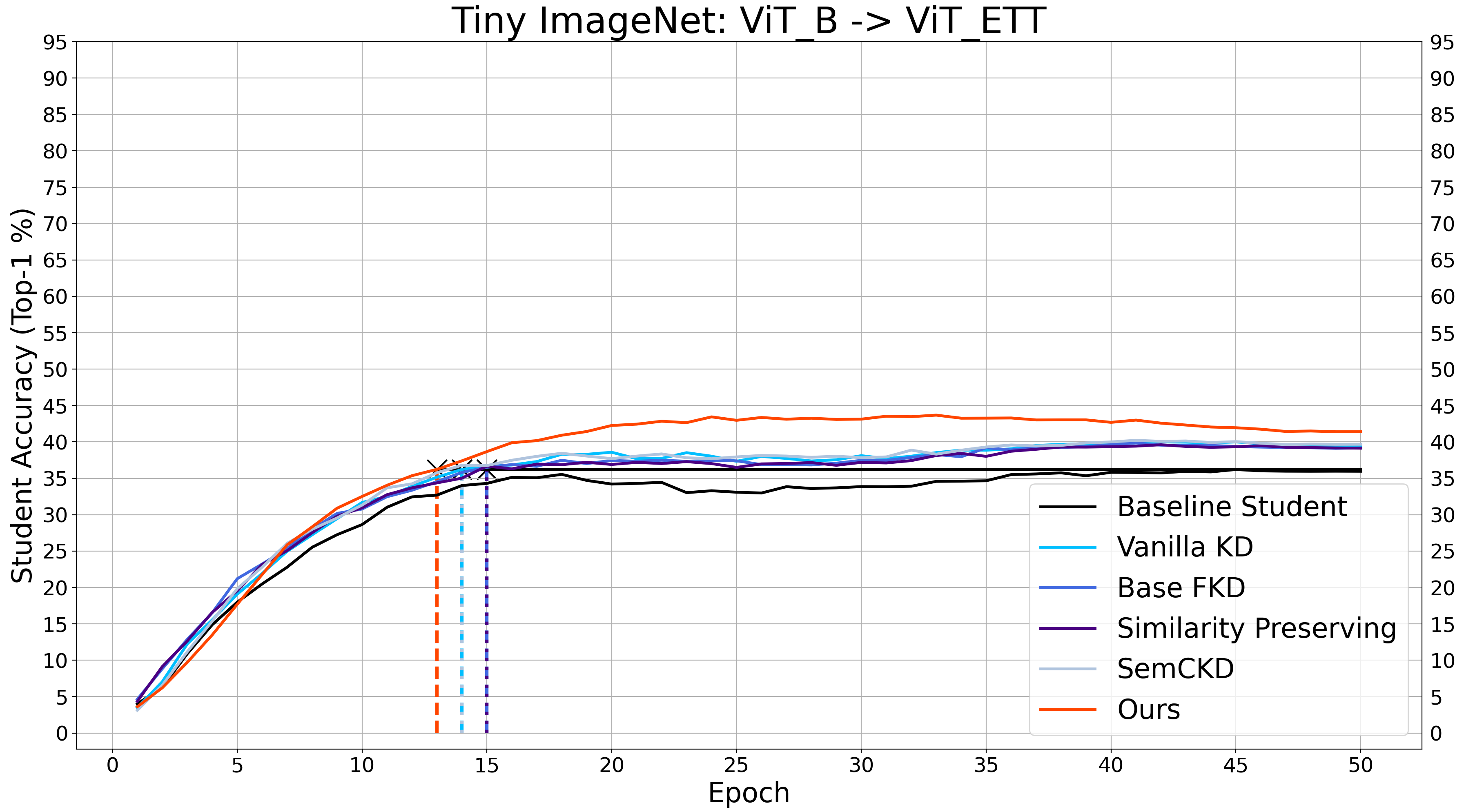}\\
    \caption{ViT\_B $\rightarrow$ ViT\_ET Training Curves. X-axis: epochs. Y-axis: top-1 accuracy. Our method shown in orange. Vertical lines indicate when each distillation method surpasses the baseline student's accuracy.}
    \label{fig:ViTCurves}
\end{figure}

We provide accuracy versus epoch curves for all model pairs presented in Experiment 1 of the main paper. Horizontal black lines are drawn at the baseline student's peak accuracy. For legibility, we plot a representative subset of baselines against the original student (without any distillation) and our method. Vertical lines are drawn at the first epoch where the student surpasses the baseline (no KD) accuracy. Results are shown in (\textbf{Figures \ref{fig:VGG11Curves}, \ref{fig:ResNet9Curves}, \ref{fig:MN2Curves}, and \ref{fig:ViTCurves}}).

\clearpage

\clearpage

\subsection{Knowledge Quality Plots}
\begin{figure}[h]
    \centering
    \includegraphics[scale=0.085]{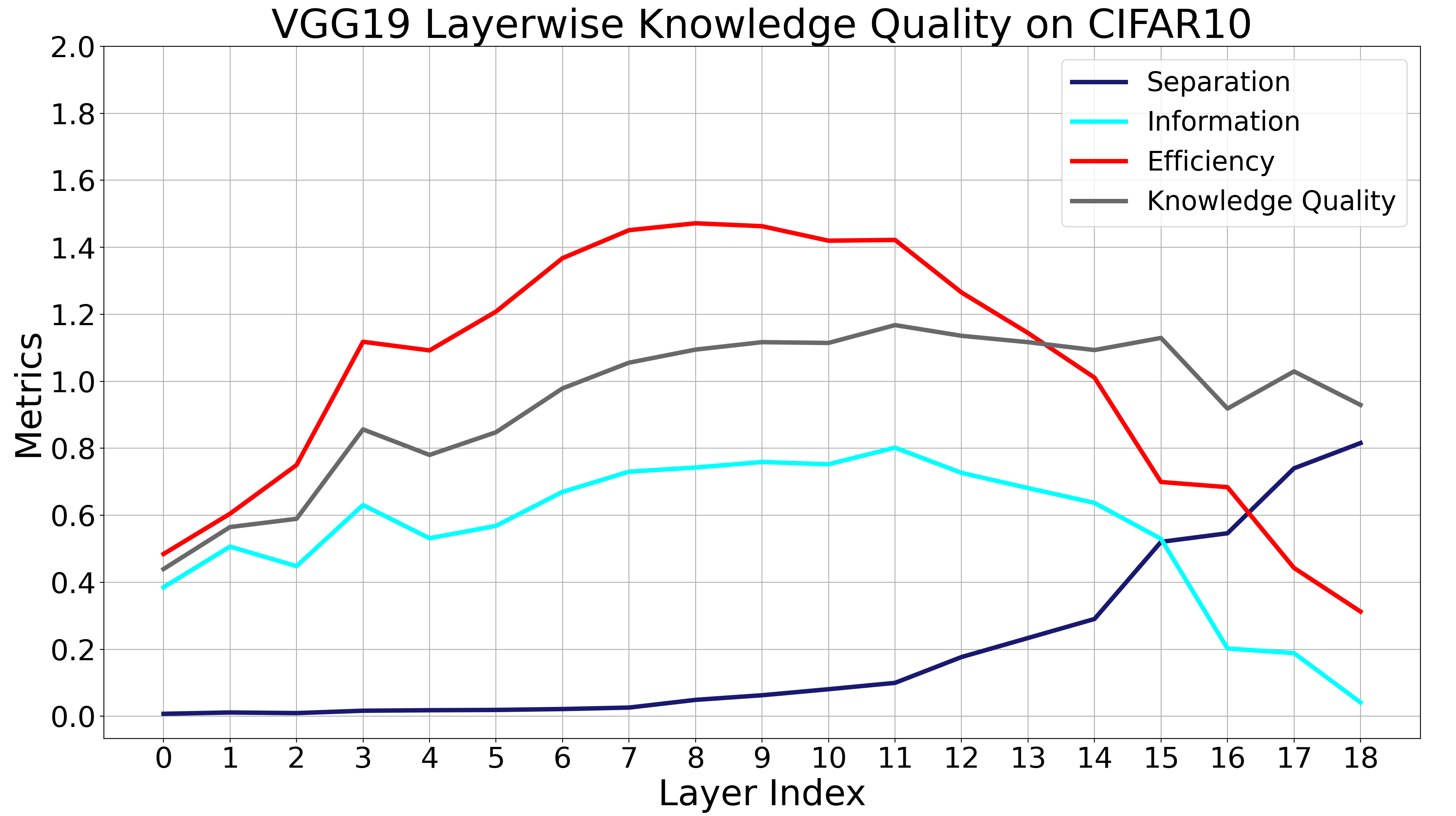}\\
    \includegraphics[scale=0.085]{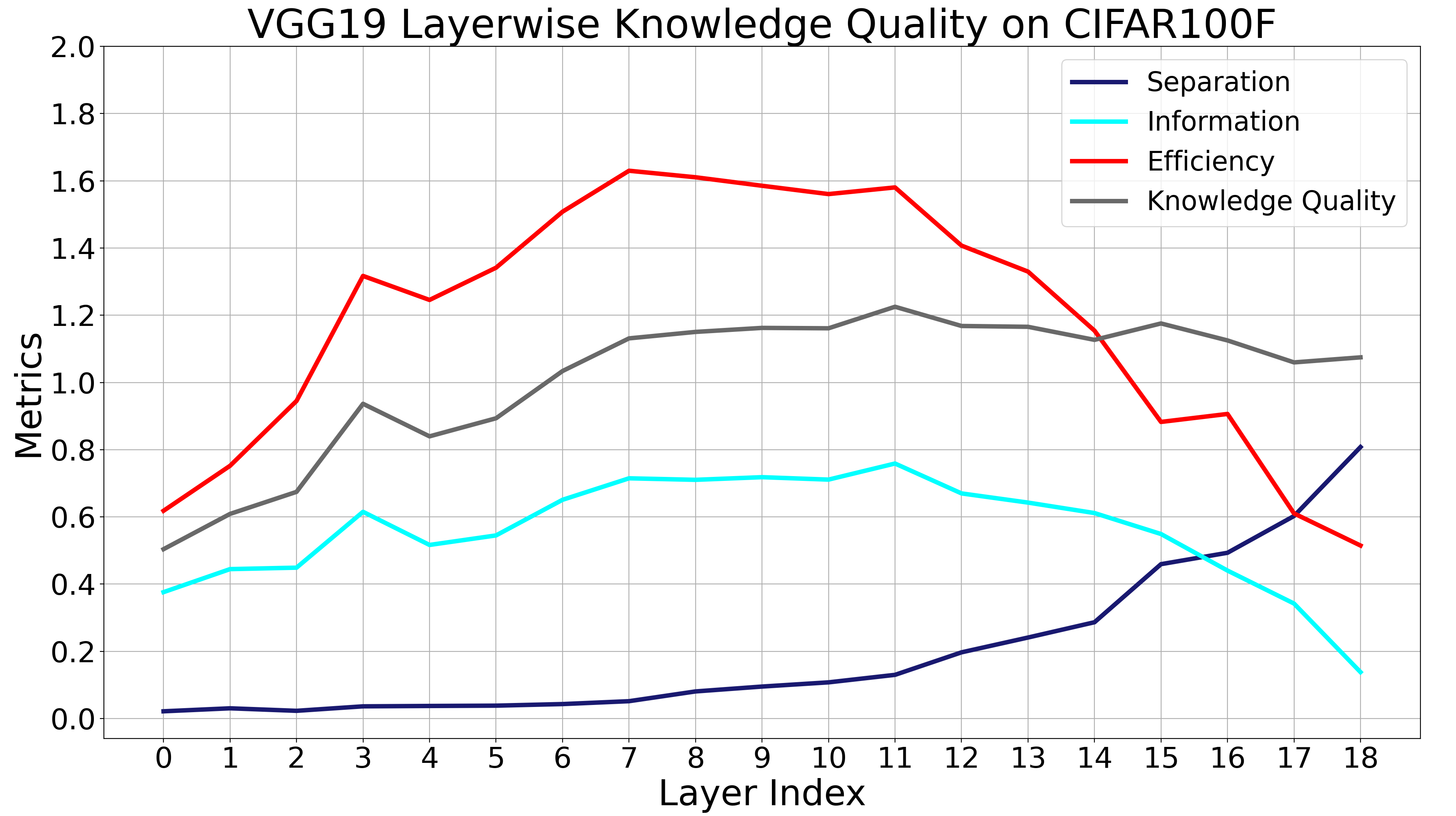}\\
    \includegraphics[scale=0.085]{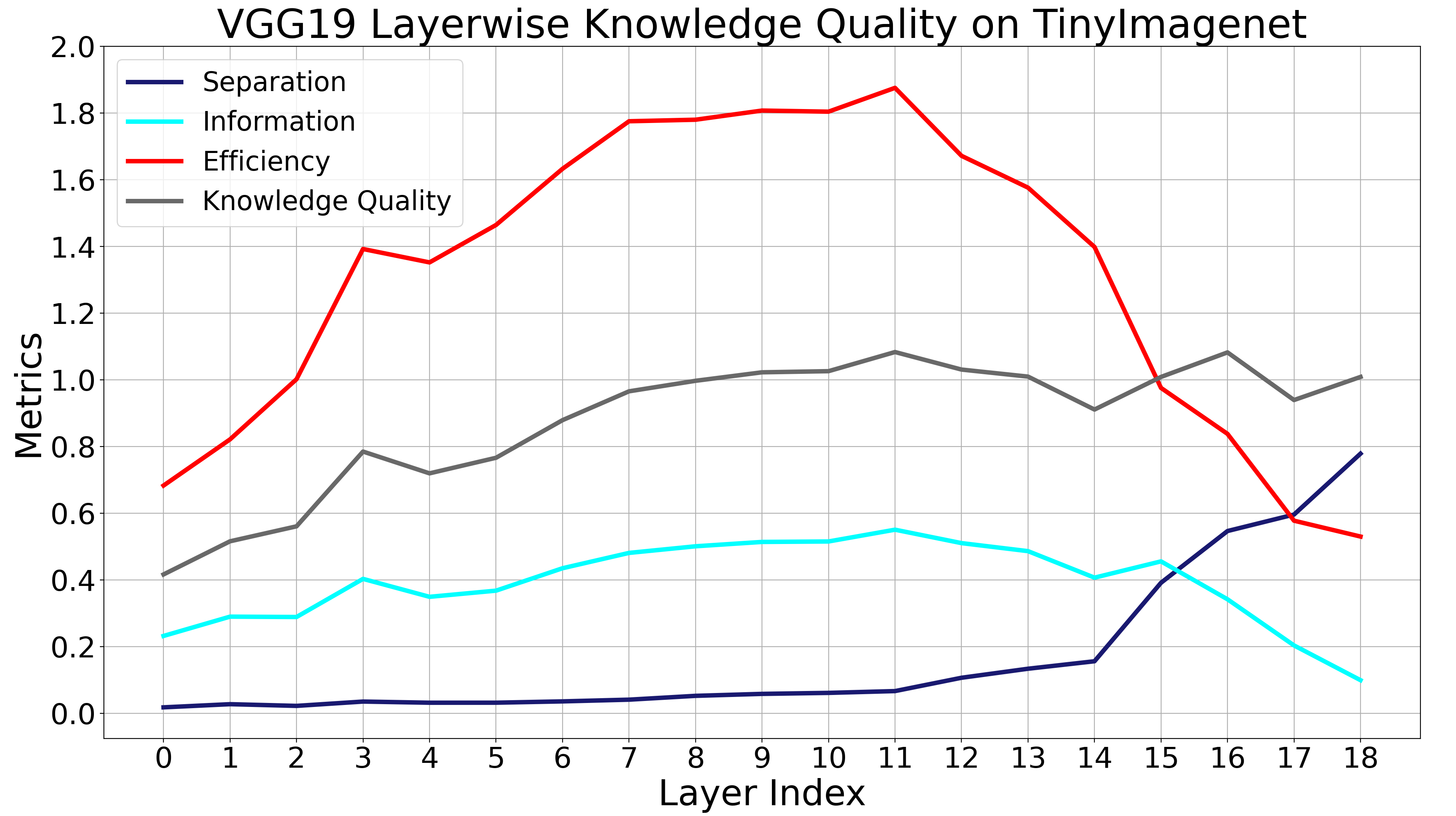}\\
    \caption{Breakdown of VGG19 Knowledge Quality. From top to bottom: CIFAR10, CIFAR100, Tiny ImageNet.}
    \label{fig:VGG19KQ}
\end{figure}

\begin{figure}
    \centering
    \includegraphics[scale=0.085]{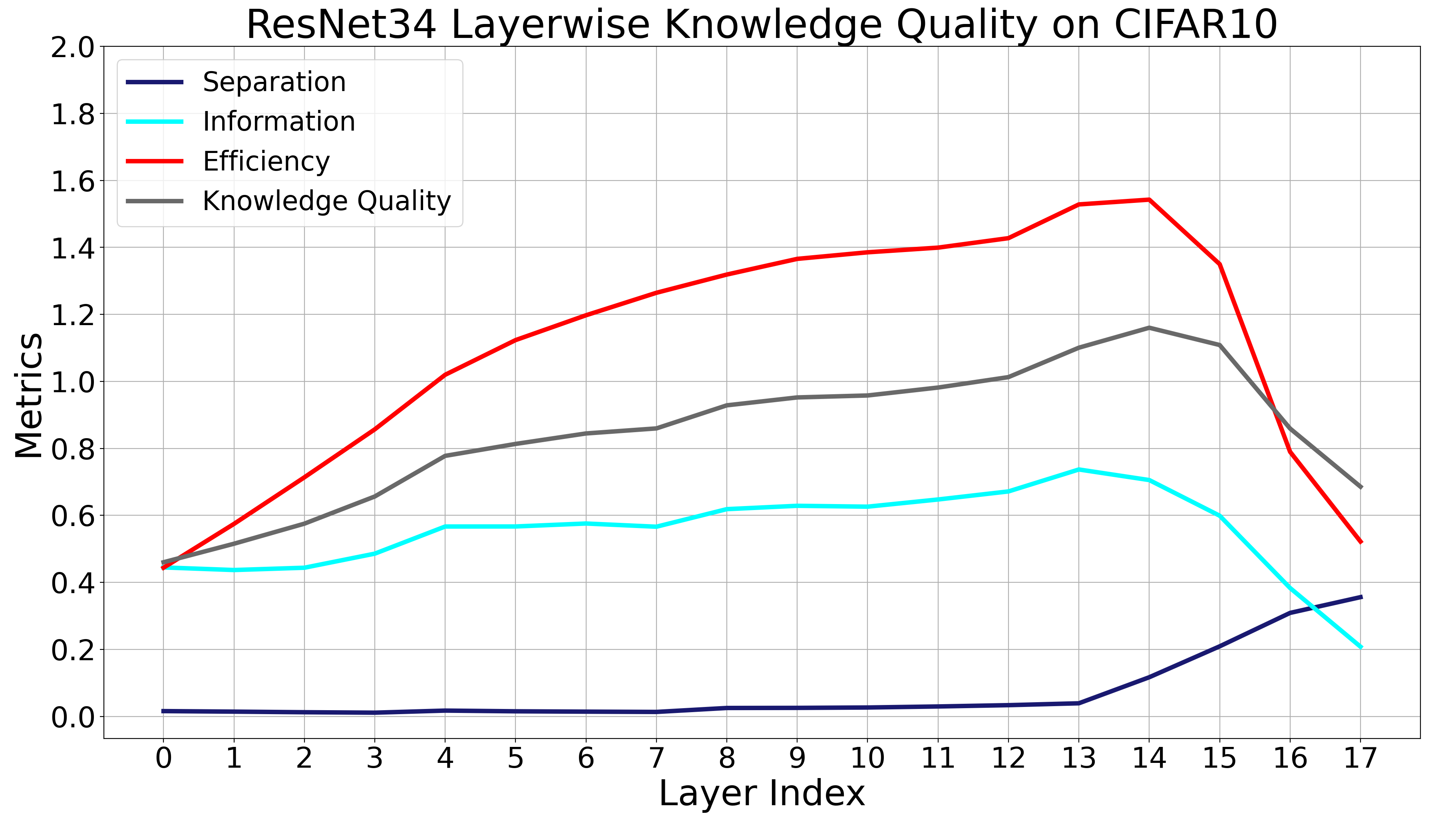}\\
    \includegraphics[scale=0.085]{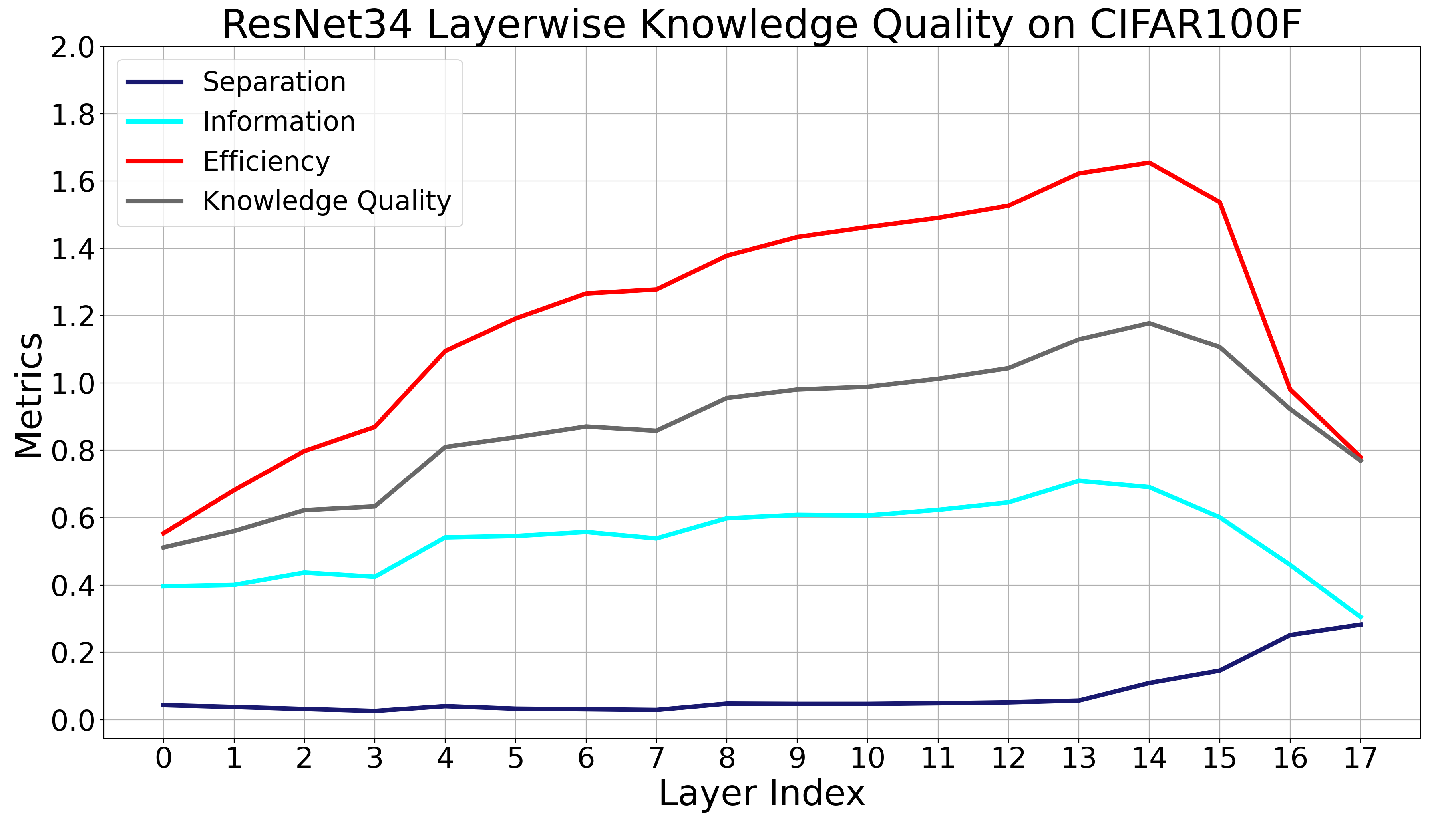}\\
    \includegraphics[scale=0.085]{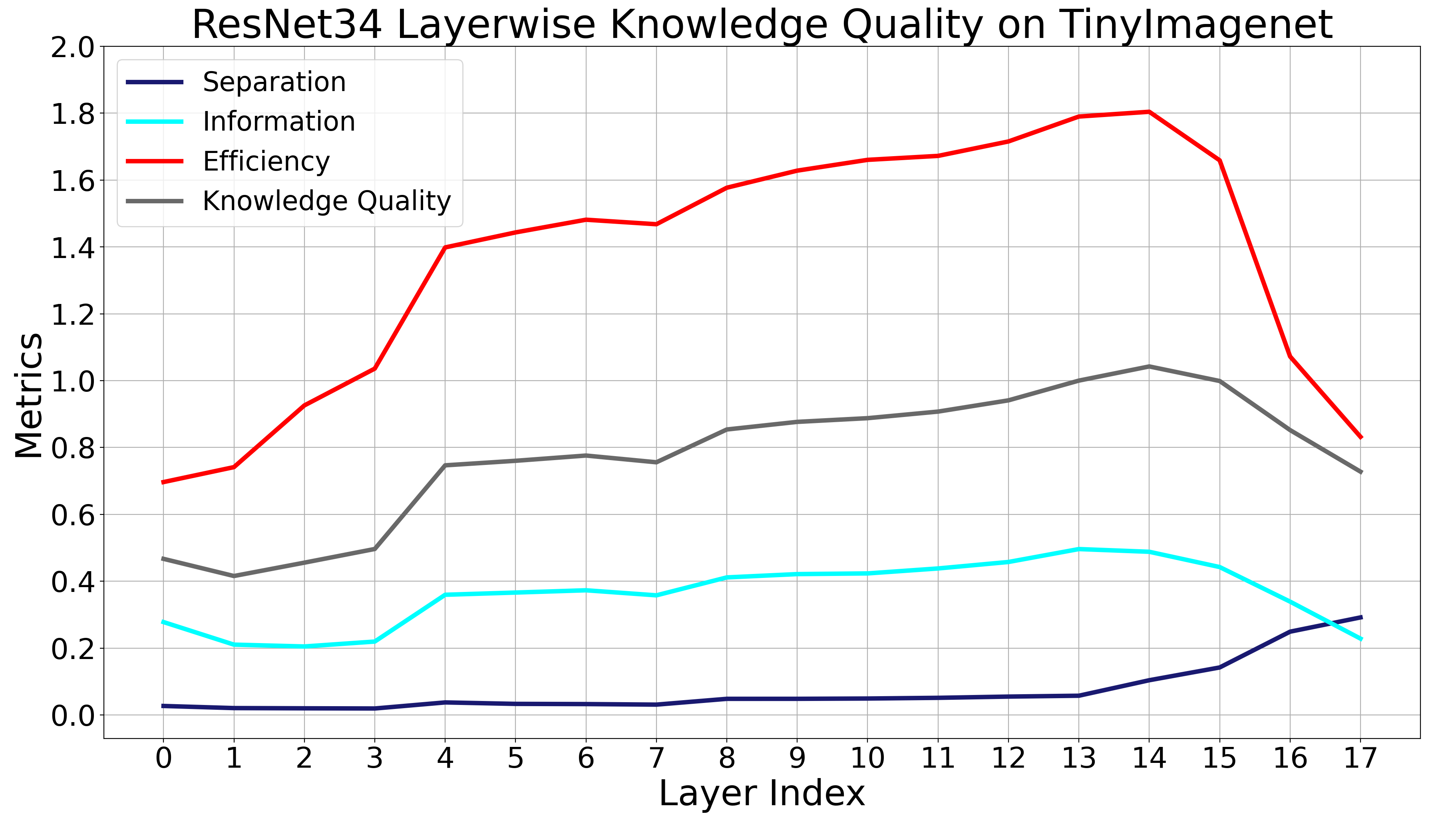}\\
    \caption{Breakdown of ResNet34 Knowledge Quality. From top to bottom: CIFAR10, CIFAR100, Tiny ImageNet.}
    \label{fig:R34KQ}
\end{figure}

\begin{figure}
    \centering
    \includegraphics[scale=0.085]{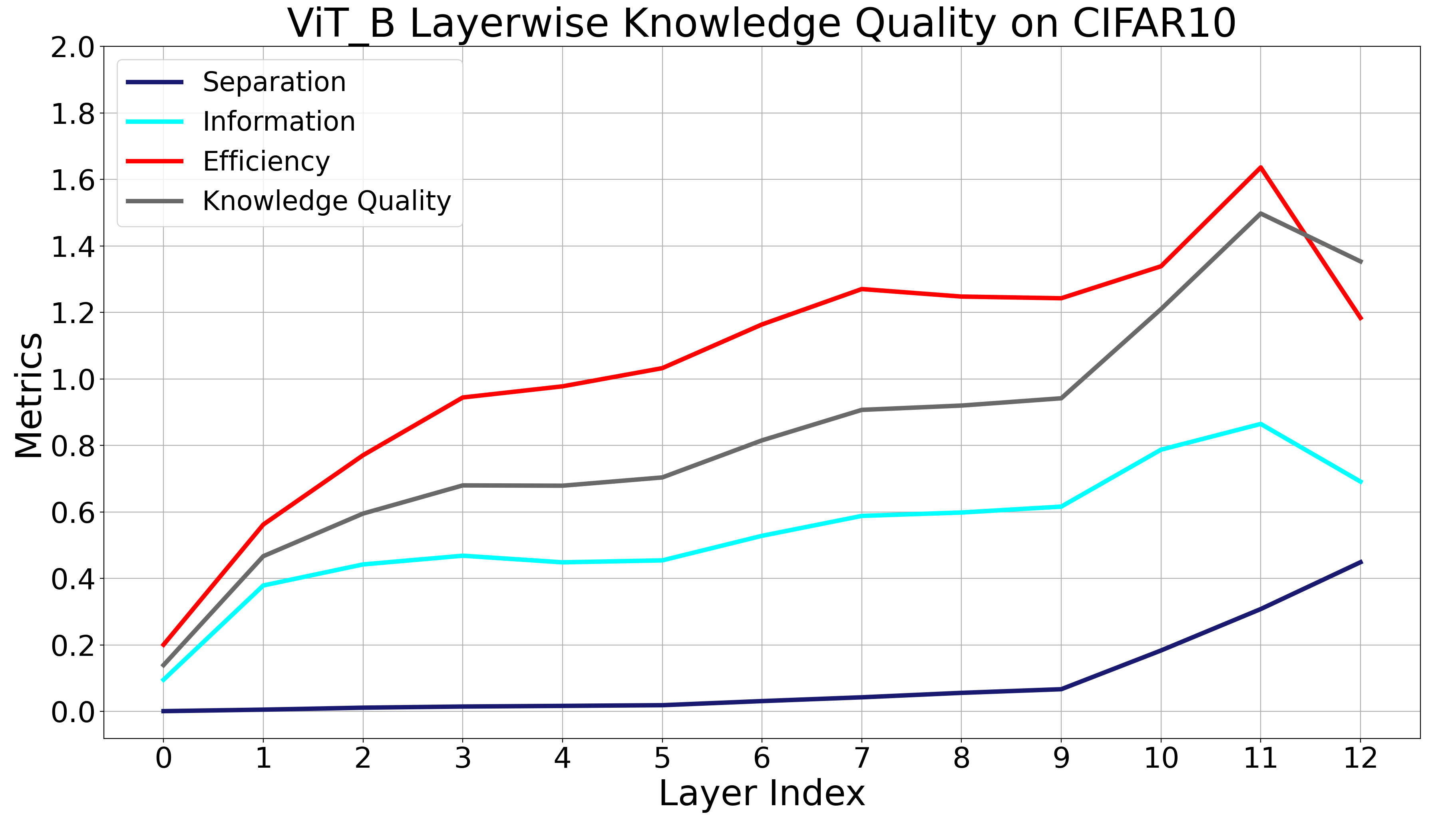}\\
    \includegraphics[scale=0.085]{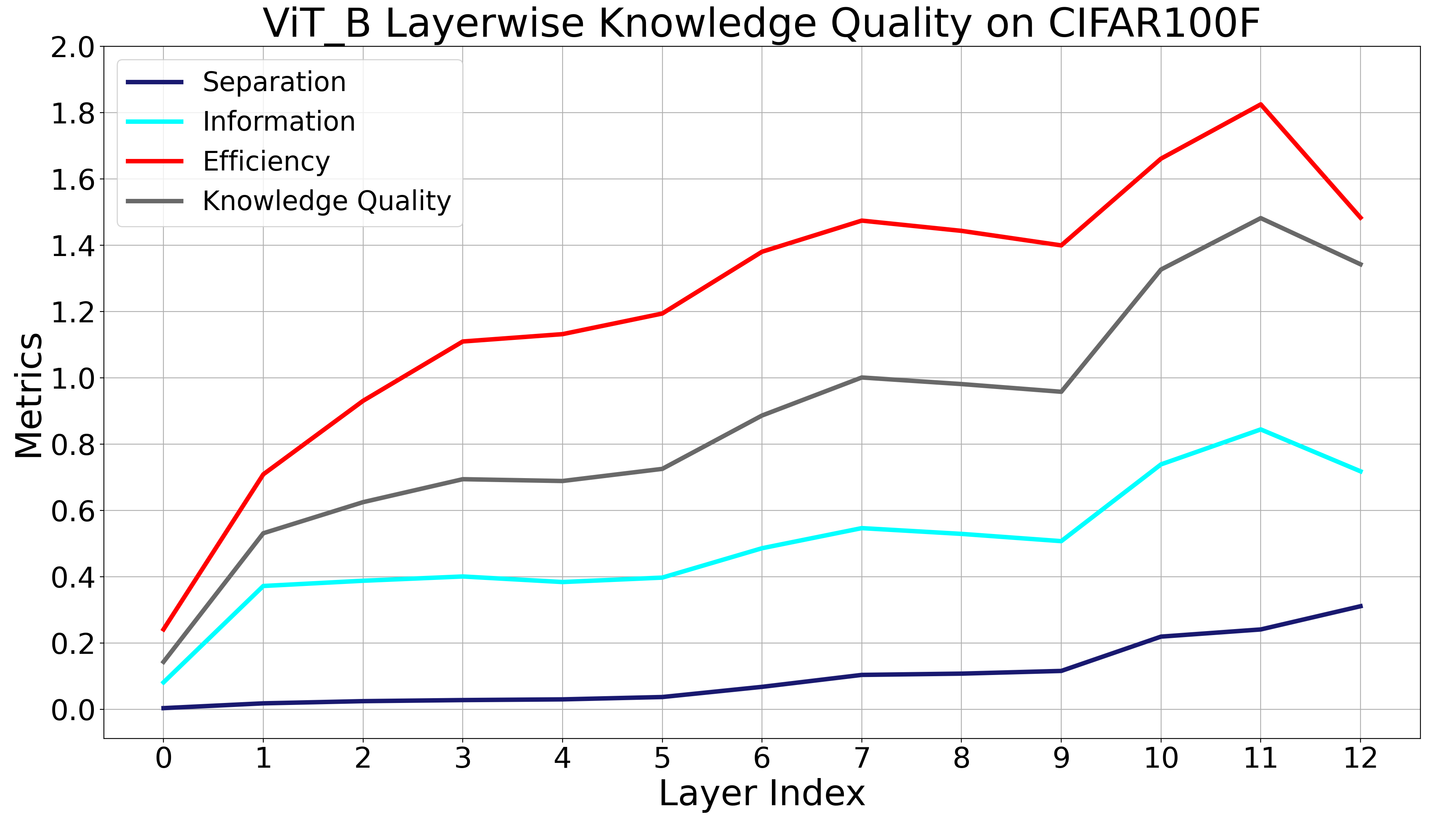}\\
    \includegraphics[scale=0.085]{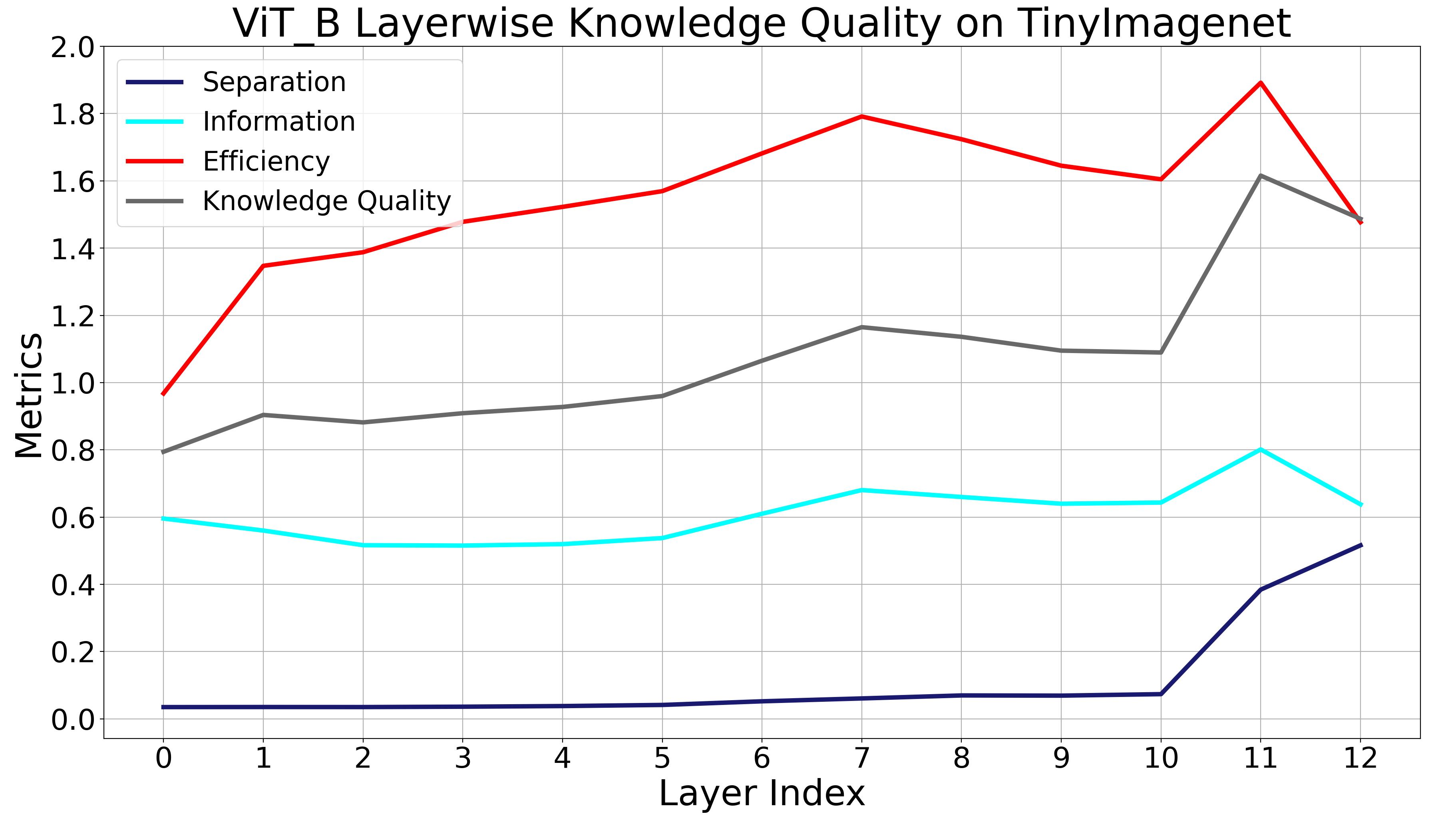}\\
    \caption{Breakdown of ViT\_B Knowledge Quality. From top to bottom: CIFAR10, CIFAR100, Tiny ImageNet.}
    \label{fig:ViTKQ}
\end{figure}

We present plots for all teachers and datasets. Per-layer curves of $\mathcal{S}, \mathcal{I}, \mathcal{E}$ are provided for all dataset-teacher combinations (\textbf{Figures \ref{fig:VGG19KQ} \ref{fig:R34KQ} \ref{fig:ViTKQ}}). As noted before, none of these curves were obtained with data augmentation to ensure reproducibility. Results begin on the next page.

\clearpage

\subsection{Teacher Layer Selections}
\subsubsection{Index Translation Tables}
As discussed in the main paper, we consider only activated representations because of the geometric implications. This results in layer indices which may not have immediately obvious relationships to the traditional definition of ``layer". Hence, we provide translation tables. They can be used to infer which ``layers" (in the traditional sense) were selected based on the knowledge quality plots and other tables to come. Note that many of the layer types listed below end in non-linear activation; e.g. residual blocks and transformer layers. We do not subdivide either of these. Flatten layers are included when they occur in activated space. For the CNNs, we also indicate the stages each layer belongs to. \textbf{Tables \ref{tab:R34ViTLayerTranslation}, \ref{tab:VGG19LayerTranslation}} contain this information.

\begin{table}[h]
\centering
\begin{tabular}{|c|c||c|}
    \hline
    Layer Type & Stage & Index\\
    \hline\hline
    \multicolumn{3}{|c|}{ResNet34}\\
    \hline\hline
    Conv2D & 0 & \\
    BatchNorm2D & 0 & \\
    ReLU & 0 & 0\\
    \hline\hline
    
    MaxPool2D & 1 & \\
    BasicBlock & 1 & 1\\
    \hline
    BasicBlock & 1 & 2\\
    \hline
    BasicBlock & 1 & 3\\
    \hline\hline

    BasicBlock & 2 & 4\\
    \hline
    BasicBlock & 2 & 5\\
    \hline
    BasicBlock & 2 & 6\\
    \hline
    BasicBlock & 2 & 7\\
    \hline\hline

    BasicBlock & 3 & 8\\
    \hline
    BasicBlock & 3 & 9\\
    \hline
    BasicBlock & 3 & 10\\
    \hline
    BasicBlock & 3 & 11\\
    \hline
    BasicBlock & 3 & 12\\
    \hline
    BasicBlock & 3 & 13\\
    \hline\hline

    BasicBlock & 4 & 14\\
    \hline
    BasicBlock & 4 & 15\\
    \hline
    BasicBlock & 4 & 16\\
    \hline\hline

    AdaptAvgPool2D & 5 & \\
    Flatten & 5 & 17\\
    \hline\hline

    Linear & 5 & \\
    \hline
\end{tabular}
\quad
\begin{tabular}{|c||c|}
    \hline
    Layer Type & Index\\
    \hline\hline
    \multicolumn{2}{|c|}{ViT\_B}\\
    \hline\hline
    ViT Input & \\
    Pos. Embedding & \\
    Dropout & \\
    EncoderBlock & 0\\
    \hline
    EncoderBlock & 1\\
    \hline
    EncoderBlock & 2\\
    \hline
    EncoderBlock & 3\\
    \hline
    EncoderBlock & 4\\
    \hline
    EncoderBlock & 5\\
    \hline
    EncoderBlock & 6\\
    \hline
    EncoderBlock & 7\\
    \hline
    EncoderBlock & 8\\
    \hline
    EncoderBlock & 9\\
    \hline
    EncoderBlock & 10\\
    \hline
    EncoderBlock & 11\\
    \hline
    LayerNorm & \\
    Flatten & 12\\
    \hline\hline
    Linear & \\
    \hline
\end{tabular}
\caption{ResNet34 (left) and ViT\_B (right) Layer Translation Tables. Stage indicates representation spatial resolution. Index denotes the layer indexing system throughout this paper.}
\label{tab:R34ViTLayerTranslation}
\end{table}

\clearpage

\begin{table}[h!]
\centering
\begin{tabular}{|c|c||c|}
    \hline
    Layer Type & Stage & Index\\
    \hline\hline
    \multicolumn{3}{|c|}{VGG19}\\
    \hline\hline
    Conv2D & 0 & \\
    BatchNorm2D & 0 & \\
    ReLU & 0 & 0\\
    \hline
    Conv2D & 0 & \\
    BatchNorm2D & 0 & \\
    ReLU & 0 & 1\\
    \hline\hline
    
    MaxPool2D & 1 & \\
    Conv2D & 1 & \\
    BatchNorm2D & 1 & \\
    ReLU & 1 & 2\\
    \hline
    Conv2D & 1 & \\
    BatchNorm2D & 1 & \\
    ReLU & 1 & 3\\
    \hline\hline
    
    MaxPool2D & 2 & \\
    Conv2D & 2 & \\
    BatchNorm2D & 2 & \\
    ReLU & 2 & 4\\
    \hline
    Conv2D & 2 & \\
    BatchNorm2D & 2 & \\
    ReLU & 2 & 5\\
    \hline
    Conv2D & 2 & \\
    BatchNorm2D & 2 & \\
    ReLU & 2 & 6\\
    \hline
    Conv2D & 2 & \\
    BatchNorm2D & 2 & \\
    ReLU & 2 & 7\\
    \hline\hline

    MaxPool2D & 3 & \\
    Conv2D & 3 & \\
    BatchNorm2D & 3 & \\
    ReLU & 3 & 8\\
    \hline
    Conv2D & 3 & \\
    BatchNorm2D & 3 & \\
    ReLU & 3 & 9\\
    \hline
\end{tabular}
\quad
\begin{tabular}{|c|c||c|}
    \hline
    Layer Type & Stage & Index\\
    \hline\hline
    \multicolumn{3}{|c|}{VGG19 Cont.}\\
    \hline\hline
    Conv2D & 3 & \\
    BatchNorm2D & 3 & \\
    ReLU & 3 & 10\\
    \hline
    Conv2D & 3 & \\
    BatchNorm2D & 3 & \\
    ReLU & 3 & 11\\
    \hline
    MaxPool2D & 4 & \\
    Conv2D & 4 & \\
    BatchNorm2D & 4 & \\
    ReLU & 4 & 12\\
    \hline
    Conv2D & 4 & \\
    BatchNorm2D & 4 & \\
    ReLU & 4 & 13\\
    \hline
    Conv2D & 4 & \\
    BatchNorm2D & 4 & \\
    ReLU & 4 & 14\\
    \hline
    Conv2D & 4 & \\
    BatchNorm2D & 4 & \\
    ReLU & 4 & 15\\
    \hline\hline
    
    MaxPool2D & 5 & \\
    AdaptAvgPool2D & 5 & \\
    Flatten & 5 & 16\\
    \hline
    Linear & 5 & \\
    ReLU & 5 & 17\\
    \hline
    Dropout & 5 & \\
    Linear & 5 & \\
    ReLU & 5 & 18\\
    \hline\hline
    Dropout & 5 & \\
    Linear & 5 & \\
    \hline
\end{tabular}

\caption{VGG19 Layer Translation Table. Stage indicates representation spatial resolution. Index denotes the layer indexing system throughout this paper.}
\label{tab:VGG19LayerTranslation}
\end{table}

\clearpage

\subsubsection{Table of Selected Indices}
\textbf{Table \ref{tab:AllLT}} contains all selected teacher layer indices $L^T$ for all selection strategies used in the main paper.

\begin{table}[h]
    \centering
    \begin{tabular}{c|c|c|c}
        $L^T$ & CIFAR10 & CIFAR100 & Tiny ImageNet\\
        \hline\hline
        \multicolumn{4}{|c|}{VGG19}\\
        \hline\hline
        Std & [3, 7, 11, 15] & [3, 7, 11, 15] & [3, 7, 11, 15]\\
        Ours & [11, 12, 13, 15] & [11, 12, 13, 15] & [11, 12, 13, 16]\\
        $\mathcal{Q} = \mathcal{S}$ & [15, 16, 17, 18] & [15, 16, 17, 18] & [15, 16, 17, 18]\\
        $\mathcal{Q} = \mathcal{I}$ & [8, 9, 10, 11] & [8, 9, 10, 11] & [9, 10, 11, 12]\\
        $\mathcal{Q} = \mathcal{E}$ & [7, 8, 9, 10] & [7, 8, 9, 11] & [8, 9, 10, 11]\\
        $\mathcal{Q} = \mathcal{IE}$ & [7, 8, 9, 11] & [7, 8, 9, 11] & [8, 9, 10, 11]\\
        \hline\hline
        \multicolumn{4}{|c|}{ResNet34}\\
        \hline\hline
        Std & [3, 7, 13, 16] & [3, 7, 13, 16] & [3, 7, 13, 16]\\
        Ours & [12, 13, 14, 15] & [12, 13, 14, 15] & [12, 13, 14, 15]\\
        $\mathcal{Q} = \mathcal{S}$ & [14, 15, 16, 17] & [14, 15, 16, 17] & [14, 15, 16, 17]\\
        $\mathcal{Q} = \mathcal{I}$ & [11, 12, 13, 14] & [11, 12, 13, 14] & [11, 12, 13, 14]\\
        $\mathcal{Q} = \mathcal{E}$ & [11, 12, 13, 14] & [12, 13, 14, 15] & [11, 12, 13, 14]\\
        $\mathcal{Q} = \mathcal{IE}$ & [11, 12, 13, 14] & [11, 12, 13, 14] & [12, 13, 14, 15]\\
        \hline\hline
        \multicolumn{4}{|c|}{ViT\_B}\\
        \hline\hline
        Std & [2, 4, 8, 10] & [2, 4, 8, 10] & [2, 4, 8, 10]\\
        Ours & [9, 10, 11, 12] & [7, 10, 11, 12] & [7, 8, 11, 12]\\
        $\mathcal{Q} = \mathcal{S}$ & [9, 10, 11, 12] & [9, 10, 11, 12] & [9, 10, 11, 12]\\
        $\mathcal{Q} = \mathcal{I}$ & [9, 10, 11, 12] & [7, 10, 11, 12] & [7, 8, 10, 11]\\
        $\mathcal{Q} = \mathcal{E}$ & [7, 8, 10, 11] & [7, 10, 11, 12] & [6, 7, 8, 11]\\
        $\mathcal{Q} = \mathcal{IE}$ & [9, 10, 11, 12] & [7, 10, 11, 12] & [7, 8, 9, 11]\\
        \hline\hline
    \end{tabular}
    \caption{Indices resulting from all teacher layer selection strategies used in the main paper. Indices can be interpreted via. the layer translation tables.}
    \label{tab:AllLT}
\end{table}

\end{document}